\documentclass[letterpaper]{article} 
\usepackage[submission]{aaai24}  
\usepackage{times}  
\usepackage{helvet}  
\usepackage{courier}  
\usepackage[hyphens]{url}  
\usepackage{graphicx} 
\usepackage{natbib}  
\usepackage{caption} 
\usepackage{subcaption}
\usepackage{arydshln}
\usepackage{url}
\usepackage[inline]{enumitem}
\usepackage{xcolor}
\usepackage{amssymb}
\usepackage{amsmath}

\usepackage{algorithm}
\usepackage{algorithmic}

\usepackage{newfloat}
\usepackage{listings}
\DeclareCaptionStyle{ruled}{labelfont=normalfont,labelsep=colon,strut=off} 
\floatstyle{ruled}
\newfloat{listing}{tb}{lst}{}
\floatname{listing}{Listing}
\title{\proposedautencoder: An Auto-encoder-based Loss Landscape Visualization Method}

\author {
    Mohannad Elhamod\textsuperscript{\rm 1},
    Anuj Karpatne\textsuperscript{\rm 1}
}
\affiliations {
    \textsuperscript{\rm 1}Virginia Tech\\
    elhamod@vt.edu, karpatne@vt.edu,
}

\usepackage{bibentry}

\usepackage[capitalise]{cleveref}
\crefname{figure}{Figure}{Figures}
\Crefname{figure}{Figure}{Figures}
\crefname{equation}{Equation}{Equations}
\Crefname{equation}{Equation}{Equations}
\crefname{section}{Section}{Sections}
\Crefname{section}{Section}{Sections}

\begin{document}

\newcommand{\proposedautencoder}{\textit{Neuro-Visualizer}}
\newcommand{\grid}{\mathcal{G}}
\newcommand{\manifold}{\mathcal{L}}
\newcommand{\trajectorymodels}{\mathcal{M_T}}
\newcommand{\gridpoints}{\mathcal{M_G}}
\newcommand{\trajectory}{\mathcal{T}}
\newcommand{\proposedautencodermodel}{\mathcal{N}}
\newcommand{\encoderlandscape}{\mathit{E}_\proposedautencodermodel}
\newcommand{\decoderlandscape}{\mathit{D}_\proposedautencodermodel}
\newcommand{\SGML}{\emph{KGML}}
\newcommand{\Lmse}{\mathrm{L_{phy\text{-}MSE}}}
\newcommand{\Ltri}{\mathrm{L_{phy\text{-}TRI}}}
\newcommand{\cophy}{\emph{CoPhy}-PGNN}
\newcommand{\PGNN}{\emph{PGNN}}
\newcommand{\eigenvec}{\boldsymbol{y}}
\newcommand{\eigenval}{b}
\newcommand{\eigenmat}{\hat{A}}
\newcommand{\nsloss}{$C$-Loss{}}
\newcommand{\eloss}{$S$-Loss}
\newcommand{\lfmodel}{\emph{CoPhy}-PGNN (Label-free)}
\newcommand{\nexmodel}{\emph{CoPhy}-PGNN (only-$\mathcal{D}_{Tr}$)}
\newcommand{\nn}{Black-box Neural Network}
\newcommand{\ncmodel}{PGNN-\emph{analogue}}
\newcommand{\BB}{black\text{-}box}
\newcommand{\lres}{\text{L}_{r}}
\newcommand{\lic}{\text{L}_{ic}}
\newcommand{\lbc}{\text{L}_{bc}}
\newcommand{\lrec}{\text{L}_{rec}}
\newcommand{\lanch}{\text{L}_{anch}}
\newcommand{\ltraj}{\text{L}_{traj}}
\newcommand{\lgrid}{\text{L}_{grid}}
\newcommand{\crec}{\text{c}_{rec}}
\newcommand{\canch}{\text{c}_{anch}}
\newcommand{\ctraj}{\text{c}_{traj}}
\newcommand{\cgrid}{\text{c}_{grid}}
\newcommand{\cres}{\text{c}_{r}}
\newcommand{\cic}{\text{c}_{ic}}
\newcommand{\cbc}{\text{c}_{bc}}
\newcommand{\ltest}{\text{L}_{test}}
\newcommand{\rlw}{\textit{RLW}}
\newcommand{\cw}{\textit{CW}}
\newcommand{\ew}{\textit{EW}}
\newcommand{\dwa}{\textit{DWA}}
\newcommand{\lranneal}{\textit{LR}$_{\textit{annealing}}$}
\newcommand{\gradnorm}{\textit{GradNorm}}
\newcommand{\lmax}{l^{max}}

\makeatletter 
\def\showauthors@on{T}
\makeatother 

\maketitle

\begin{abstract}
In recent years, there has been a growing interest in visualizing the loss landscape of neural networks. Linear landscape visualization methods, such as principal component analysis, have become widely used as they intuitively help researchers study neural networks and their training process. However, these linear methods suffer from limitations and drawbacks due to their lack of flexibility and low fidelity at representing the  high dimensional landscape. In this paper, we present a novel auto-encoder-based non-linear landscape visualization method called \proposedautencoder{} that addresses these shortcoming and provides useful insights about neural network loss landscapes. To demonstrate its potential, we run experiments on a variety of problems in two separate applications of knowledge-guided machine learning (KGML). Our findings show that \proposedautencoder{} outperforms other linear and non-linear baselines and helps corroborate, and sometime challenge, claims proposed by machine learning community. All code and data used in the experiments of this paper are available at an anonymous link\footnote{\url{https://anonymous.4open.science/r/NeuroVisualizer-FDD6}}.
\end{abstract}


 \section{Introduction} \label{se:lit_trainExplainability}

    
            
            Understanding the loss landscape of deep neural network has attracted much attention in recent years, both from theoretical and visualization standpoints. 
            In this work, we focus on the problem of \textit{loss landscape visualization}, which is the practice of plotting a neural network's loss function w.r.t its high-dimensional model parameters (i.e., weights and biases) on a low-dimensional embedding space, usually 1-D \cite{goodfellow2014qualitatively} or 2-D \cite{NIPS2018_7875}, to qualitatively assess generalization performance and training convergence. 
            One of the seminal works that established loss landscape visualization as an investigative tool in the field is the work by \cite{NIPS2018_7875}.
            This work introduced the approach of projecting model parameters to random planes in low dimensions (usually 2D), to visualize and assess the quality of the loss surface at the vicinity of a converged model, including  whether the loss surface is ill-regularized and filled with local minima. 

            However, when it comes to visualizing multiple models (e.g., the set of models forming a training trajectory), the selection of the projection plane becomes more challenging as the chosen plane should capture ``interesting" properties of the loss landscape for the entire set of models, not just a single one. Naturally, the most common way for finding this plane is by performing a Principal Component Analysis (PCA) on the set of models, selecting the two main principal components, and visualizing the landscape on that 2D plane such that it passes through the trajectory's final model \cite{NIPS2018_7875}. 
            Another approach is to use the two eigenvectors with the highest eigenvalues \cite{chatzimichailidis2019gradvis, pyhessian}. 
   Despite how helpful these projections can be, each of these practices have their own advantages and disadvantages, as detailed below.

                First, since training trajectories do not necessarily fit on a 2-D plane, the use of linear projection methods such as PCA yield loss surfaces that are mostly accurate at the point of intersection, 
                which is generally the trajectory's final model, 
                but drop in accuracy as we move away from that point. 
                Second, while 
                some previous works on loss landscape visualization have used non-linear methods such as UMAP and t-SNE \cite{pmlr-v137-huang20a}, SHEAP \cite{PhysRevX.11.041026}, and PHATE \cite{phate}, these methods are more suitable for visualizing the relationship between models (e.g., model clustering) instead of visualizing {landscapes} of model trajectories. 
                Third, other non-linear approaches such as Locally Linear Embeddings (LLE) \cite{doi:10.1126/science.290.5500.2323}, and Laplacian Eigen-maps \cite{Wang2012} suffer from the lack of an inverse transform, 
                making them unsuitable for landscape visualization.
                    A fourth issue with landscape visualization methods is the scale at which the loss surface is visualized. While the most common practice is to use `filter normalization' \cite{NIPS2018_7875} to make the visualization scale invariant, this approach only works when visualizing the loss landscape at the vicinity of a single model, and is not applicable when visualizing training trajectories. 
                
                To mitigate the aforementioned shortcomings, this paper focuses on answering the question: can we devise a non-linear projection method for loss landscape visualization that: \textbf{(1) \textit{faithfully} captures training trajectories and loss landscapes in their vicinity}, thus improving model optimization understanding, and \textbf{(2) \textit{adaptively} scales the projection space} based on problem requirements?


                

            In response to the above question, we propose a novel auto-encoder-based non-linear loss landscape visualization approach called \proposedautencoder{}. While our proposed approach can be applied to any general problem, we specifically demonstrate its potential in the context of two applications in the field of knowledge-guided machine learning (KGML) \citep{tgds}: solving the Schrodinger's equation in quantum mechanics using the framework of physics-guided neural networks with competing physics (\cophy{}) \cite{elhamod2022cophy}, and solving generic partial differential equations (PDEs)  using the framework of physics-informed neural networks (PINNs) \cite{raissi2017physics1}. While  these applications in KGML have received considerable attention in recent years, what is missing in the field is a comprehensive understanding of the effects of adding physics-guided loss functions on the loss landscape of neural networks.
            By using \proposedautencoder{}, we are able to visualize and discover novel insights about the performance of competing KGML approaches proposed for the two applications, corroborating, and in some places even challenging, optimization claims proposed in previous KGML literature. Our work thus forges a novel \textit{``collaborative bridge''} between two sub-fields of AI: neural loss landscape visualization and KGML. We anticipate our work to serve as a starting point for other researchers to develop novel visualization approaches for KGML in the future.
            
            

       \section{Related Works} \label{se:litreview}

            Though qualitative in nature, the analysis of neural loss landscapes through visualization approaches  \cite{NIPS2018_7875} is becoming a more common practice \cite{NIPS2018_8095, mei2018mean, nguyen2018loss} as an alternative to quantitative methods such as the Fisher information matrix (FIM) analysis \cite{karakida2019normalization} and Hessian analysis \cite{ma2022beyond, guiroy2019understanding}. An example use-case of using visualization tools is to determine the impact of the loss landscape structure (e.g., flatness, valleys, and basins) on model generalization and overfitting  \cite{pmlr-v137-huang20a, sypherd2020alphaloss, yang2021taxonomizing, prabhu2019understanding, xu2019understanding}.
            Other examples include understanding model optimization \cite{pmlr-v137-huang20a, ma2022beyond, keskar2017on, 9194023, yang2021taxonomizing}, assessing the generalization of Model-Agnostic Multi-task Learning (MAML) \cite{guiroy2019understanding}, investigating the smoothing effect of noise over sharp minima \cite{wen2018smoothout}, and studying the effectiveness of skip connections in removing bad valleys with sub-optimal minima during optimization \cite{nguyen2019optimization}.

  \section{Proposed Method: \proposedautencoder{}} \label{se:proposedlandscape}

            As we have discussed the drawbacks of existing loss landscape visualization methods in \cref{se:lit_trainExplainability}, it is appropriate to echo \cite{PhysRevX.11.041026}'s thoughts on finding a non-linear manifold ``such that the source data lie on, or close to, some low-dimensional manifold embedded within the original high-dimensional space". Hence, we propose using a neural auto-encoder, dubbed \proposedautencoder{}, to learn a non-linear manifold that embeds the points of interest (i.e., models) in the high-dimensional loss landscape.


            \subsection{Formal Definition}
            
            Let's assume that we have a trajectory $\trajectory$ that consists of a set of models $\trajectorymodels \subset \mathbb{R}^n$ where $\mathbb{R}^n$ is the $n$-dimensional model parameter space. We want to learn a 2-D manifold $\manifold$ such that $\trajectorymodels \subset \manifold$. This manifold is to be scaled and visualized as a grid $\grid \subset \mathbb{R}^2$. For convenience, and without the loss of generality, we standardize the grid to be strictly $\grid = [-1, +1]\times [-1, +1] $. 
            This task naturally poses itself to be solved by auto-encoders, which generally take a high-dimensional input (e.g., $\mathbb{R}^n$), and map it onto a low dimensional one (e.g., $\mathbb{R}^2$). In the process of training an auto-encoder, the training data is fit into a learned low-dimensional manifold 
            \cite{bengio2009learning}. Mathematically, we propose learning a \proposedautencoder{} auto-encoder $\proposedautencodermodel : \manifold \to \grid$ which consists of an encoder $\encoderlandscape$ and a decoder $\decoderlandscape$, such that  $z = \encoderlandscape (m \in \manifold) \in \grid$ and $m^\prime = \decoderlandscape (z \in \grid) $.

            To train the parameters of the auto-encoder $\theta_\proposedautencodermodel$, a reconstruction loss is minimized:
            \begin{equation}
                \lrec =  \text{MSE}_{\trajectorymodels} \left[ m_{i} ,  \proposedautencodermodel({m_i}) \right]
            \end{equation}
            As $\proposedautencodermodel$ gets optimized, it learns 
            a manifold that contains the training data points (e.g., the trajectory models $\trajectorymodels$).

            \subsection{Additional  Constraints in \proposedautencoder{}} \label{se:constaints}
            
            While the reconstruction loss is sufficient to guarantee learning a manifold that embeds the trajectory models, it does not guarantee any other properties of this manifold beyond continuity. However, this turns out to be a feature, not a bug, as additional desired properties of the embedding space can be imposed by adding other constraints in the form of loss functions. This is in contrast to baseline linear methods (e.g., PCA), where once the projection method is selected, there is little control over the properties of the resultant plane or manifold. 
                        Here, we list some interesting and useful constraints that we later adopt in our experiments in \cref{resultslandscape}. This list, however, is not exhaustive;
            \proposedautencoder{} is flexible and can be customized with many other possible constraints.These constraints can also be combined as a weighted sum in a multi-task learning formulation. 

            
            \subsubsection{Location anchoring constraints:} \label{p:anchoring}
            This type of constraints anchors a set of points (e.g., trajectory models) onto certain locations on the grid. This helps orient the training trajectory such that certain aspects of the optimization process are highlighted. The general form of a location anchoring constraint is:


            \begin{equation}
                \lanch = \text{MSE}_{\trajectorymodels^\prime \subseteq \trajectorymodels} \left[ \encoderlandscape(m_{i}) , \mathcal{A}_i \right],
            \end{equation}
            where $\mathcal{A} \subset \grid$ is the set of desired anchoring points on the grid and $\trajectorymodels^\prime$ is the set of models that correspond to those anchoring points.
            In this work, we chose to demonstrate three  examples of anchoring constraints as detailed below (see \cref{app:hyper} for further details): 

            \begin{itemize}[itemsep=0em]
            \item \textit{Polar pinning (${\lanch}_1$) }: This constraint places the trajectory's first and last models at the bottom left and top right corners of the grid, respectively. This helps stretch the trajectory across the grid and utilize the entire space.
            \item \textit{Center pinning (${\lanch}_2$) }: This constraint positions the last model at the center of the grid---a perspective suitable for showing the final stages of optimization in  detail.
            \item \textit{Circle pinning (${\lanch}_3$) }: This constraint positions the trajectory models' projections equally distant from each other on a circle with a specified radius.
            \end{itemize}

            
            
            

            \subsubsection{Grid scaling constraints:} \label{p:gridscaling}

            Another type of constraints can be devised to ensure the grid has a certain scale. Unlike PCA, \proposedautencoder{}'s grid does not generally have a uniform and linear scale. Rather, it is more flexible with a variable scaling factor across the grid, allowing it to show more details at certain areas while zooming out on the rest. To capitalize on this property, we construct a constraint to capture and control the zooming behavior as follows.
            We scale the grid such that points in the vicinity of trajectory models, which is of more interest and importance, has a relatively higher density than other distant points. 
            To formalize this, we construct the following grid scaling loss:

            \begin{equation}
               \lgrid = \text{MSE}_{m \in \gridpoints} \left[ \log{\left( d_m \right) - l_m , \log{\left( d^{max} \right)} - \lmax} \right]
            \end{equation}
            where $d_m$ is the distance between a grid mesh-point $m$ and the closest trajectory point to it \underline{in the parameter space}, $d^{max}$ is the distance between the first and last model on the trajectory, and $l_m$ is the distance equivalent to $d_m$ \underline{in the grid space}. Finally, $\lmax$ is a hyper-parameter chosen based on the desired scaling factor. 
            By minimizing $\lgrid$, a constant logarithmic scale between grid space and parameter space distances is enforced. The larger the value of $\lmax$, the higher the zooming effect around the trajectory.
        

            

        \section{Results and Applications} \label{resultslandscape}

While assessing the ``correctness” of  loss landscape visualization methods is non-trivial (see \cref{app:correctness}), we demonstrate the usefulness of \proposedautencoder{} in discovering novel insights and its advantages compared to existing loss landscape visualization methods in the context of two KGML applications. 

            \subsection{Applying \proposedautencoder{} on \cophy{}} \label{ss:cophy_landscape2}

            To show the effectiveness of \proposedautencoder{}, we take \cophy{} \cite{elhamod2022cophy} in quantum mechanics as a use-case application. Within the Knowledge-Guided Machine Learning (\SGML{}) framework \citep{tgds}, \cophy{} is a neural network that is trained with adaptively balanced physics loss terms in order to solve eigen-value problems with better generalization than a purely data-driven neural network (see \cref{app:cophy} for details).

            \subsubsection{Qualitative and Quantitative Assessments of \proposedautencoder{} Against Other Baselines.}
            \label{ss:proposedvsbaselines}
\begin{figure*}[htbp]
              \centering
              \begin{subfigure}[b]{0.3\textwidth}
                \includegraphics[width=\textwidth]{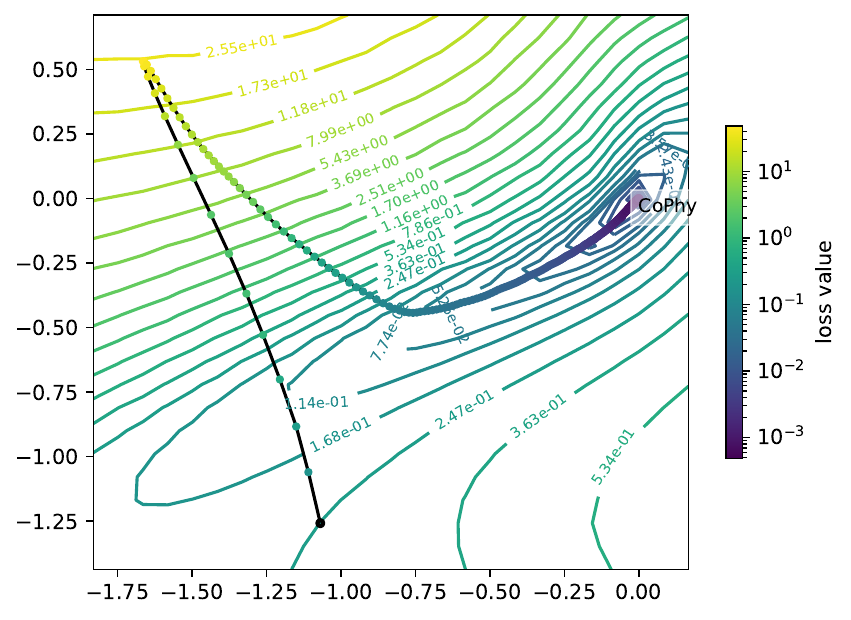}
                \caption{PCA}
                \label{fig:PCAvsAE_outPCA}
              \end{subfigure}
              \hfill
             \begin{subfigure}[b]{0.3\textwidth}
                \includegraphics[width=\textwidth]{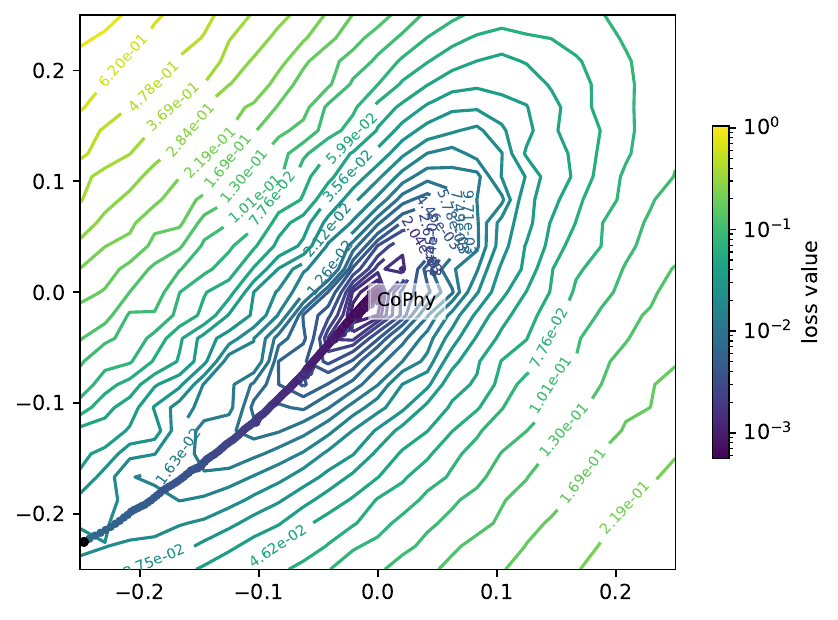}
                \caption{PCA (zoomed-in)}
                \label{fig:PCAvsAE_zoomedPCA}
              \end{subfigure}
              \hfill
              \begin{subfigure}[b]{0.3\textwidth}
                \includegraphics[width=\textwidth]{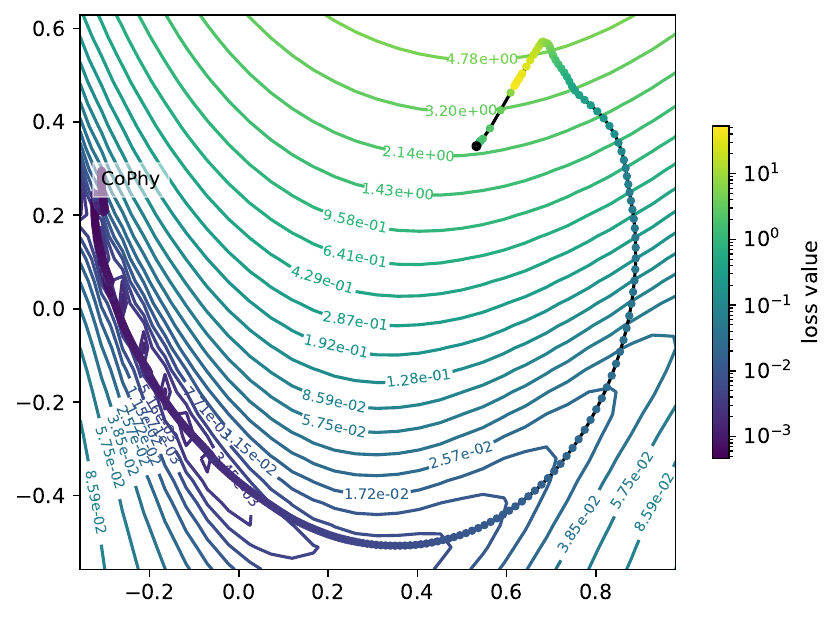}
                \caption{Kernel-PCA}
             \label{fig:KernelPCAloss}
              \end{subfigure}
              \begin{subfigure}[b]{0.3\textwidth}
                \includegraphics[width=\textwidth]{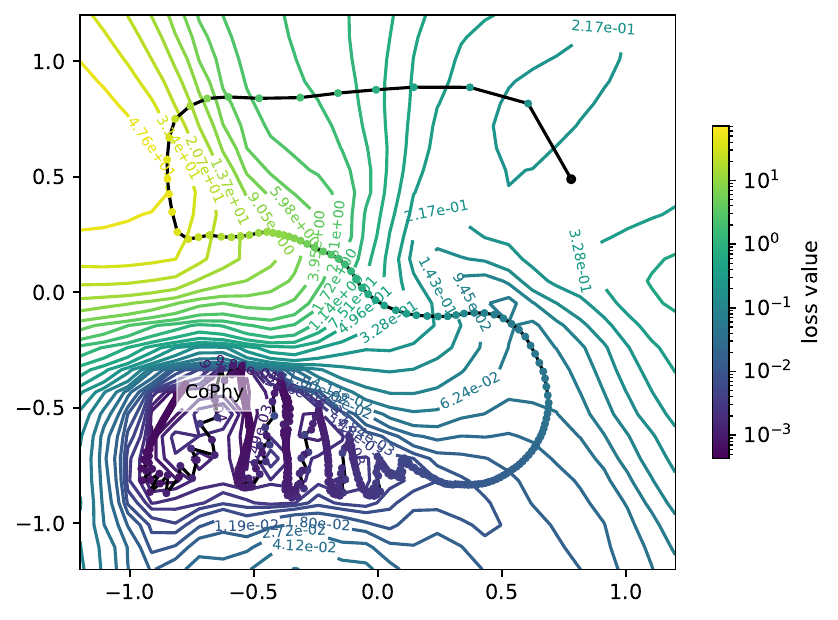}
                \caption{\proposedautencoder{}}
             \label{fig:PCAvsAE_out}
              \end{subfigure}
              \hfill
              \begin{subfigure}[b]{0.33\textwidth}
                \includegraphics[width=\textwidth]{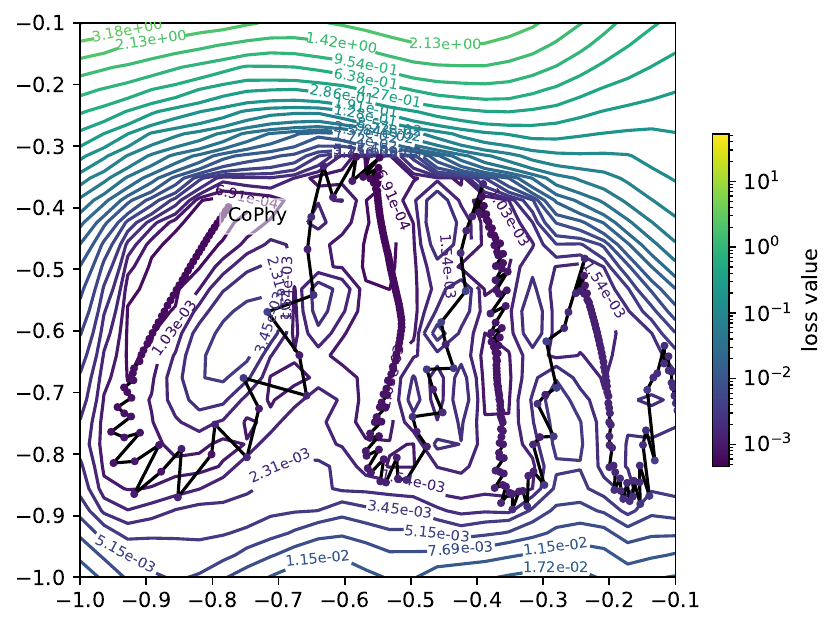}
                \caption{\proposedautencoder{} (zoomed-in)}
                \label{fig:PCAvsAE_zoomed}
              \end{subfigure}
              \hfill
                \begin{subfigure}[b]{0.3\textwidth}
                \includegraphics[width=\textwidth]{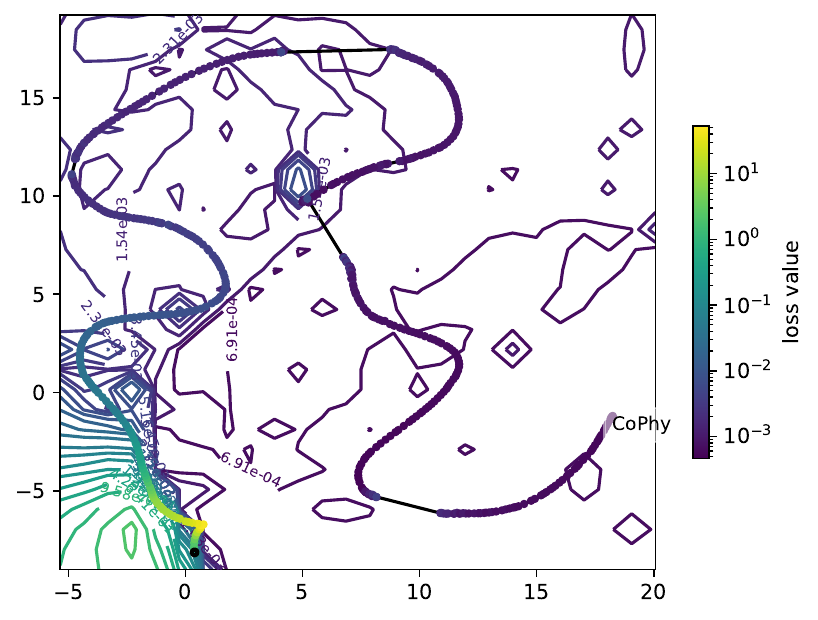}
                \caption{UMAP}
                \label{fig:UMAPloss}
              \end{subfigure}

              \caption{A comparison of \proposedautencoder{} and other baselines in terms of the consistency of \cophy{}'s overall physics loss values between trajectory models and their corresponding manifold projections. 
              Clearly, \proposedautencoder{} shows richer details and a manifold that better fits the trajectory models.}
              \label{fig:PCAvsAE}
            \end{figure*}

                        \cref{fig:PCAvsAE} compares \proposedautencoder{}'s and other baselines' overall physics loss landscapes for \cophy{}. Note that the contours in all sub-figures, except the zoomed-in ones, have been scaled equally for fair comparison. We make the following observations.

            First, \textbf{\proposedautencoder{} shows richer details.} In the top row of \cref{fig:PCAvsAE}, both PCA and \proposedautencoder{} show that, due to optimizing multiple conflicting loss terms, the training took a detour before descending into a terminal minima. \proposedautencoder{}'s story in \cref{fig:PCAvsAE_out}, however, is much richer. Here, we can see that the ``terminal" minima is not truly a simple minima, but rather a basin of a complex surface and with many local minima, causing the model to bounce around during the final stages of its optimization. This observation becomes even clearer as we zoom into the convergence area (see middle column). Being able to observe such complexity of the loss landscape is crucial for determining how well-designed the loss terms are for training.
            
            Second, \textbf{\proposedautencoder{}'s manifold fits the trajectory models better.} Looking at the model colors, which corresponds to their loss values, in \cref{fig:PCAvsAE}, we can see that the actual model loss values for PCA (\cref{fig:PCAvsAE_zoomedPCA}) do not quite match with their corresponding locations on the learned manifold (i.e., point and contour colors do not generally match at their corresponding locations). 
            The explanation for this discrepancy is that PCA is a linear method that only intersects with the high-dimensional trajectory at one point.
            This is unlike \proposedautencoder{} (\cref{fig:PCAvsAE_out,fig:PCAvsAE_zoomed}) where the points match in color with their underlying contours because the learned 2-D manifold passes through all the trajectory models, with good approximation. 

          To compare \proposedautencoder{} to other non-linear methods, we visualize the same trajectory using UMAP \cite{mcinnes2018umap} and Kernel-PCA \cite{10.1007/BFb0020217} with an RBF kernel in 
          \cref{fig:PCAvsAE}. Compared to \proposedautencoder{}, we make two observations. First, UMAP uses almost the entire grid to show the local minima at which the model arrives, with little attention to the initial stage of optimization. Thus, UMAP fails to give the full picture of the training trajectory when compared to \proposedautencoder{}. Similarly, while Kernel-PCA is a non-linear method, its landscape visualization looks too simplistic with little insight to provide beyond that of PCA's. 
                    \cref{app:errors_cophy} provides more visualizations on the projection and loss errors of \proposedautencoder{} compared to the other baselines. For a more quantitative assessment, however, we provide some comparative metrics in \cref{tab:quant_losslandscape}. By looking at both the average relative error in loss values and the average projection error in the parameter space, we can see that \proposedautencoder{} beats all the other methods by orders of magnitude.

            \begin{table}
            \centering
            \begin{tabular}{|p{1cm}|p{1.5cm}|c|p{1cm}|c|}
            \hline
            \textbf{Metric} & \proposedautencoder{}& PCA & Kernel-PCA & UMAP   \\
            \hline
            \textbf{$e_{\text{relative}}$} & 0.0095 & 1.6782  & 4.7250 & 0.4295\\
            \hline
            \textbf{$e_{\text{proj}}$} & 0.0005 &  0.2832 & 0.0865 & 0.2307 \\
            \hline
            \end{tabular}
            \caption{A quantitative comparison of \proposedautencoder{} against other baselines in terms of average relative physics loss error \textbf{$e_{\text{relative}}$} and average projection error \textbf{$e_{\text{proj}}$}. \proposedautencoder{} outperforms all other baselines across the board.}
            \label{tab:quant_losslandscape}
            \end{table}

          \subsubsection{Using \proposedautencoder{} to Study the Advantages of the \cophy{} Approach.} 
\begin{figure}[htp]
              \centering

              \begin{subfigure}[b]{0.3\textwidth}
                \includegraphics[width=\textwidth]{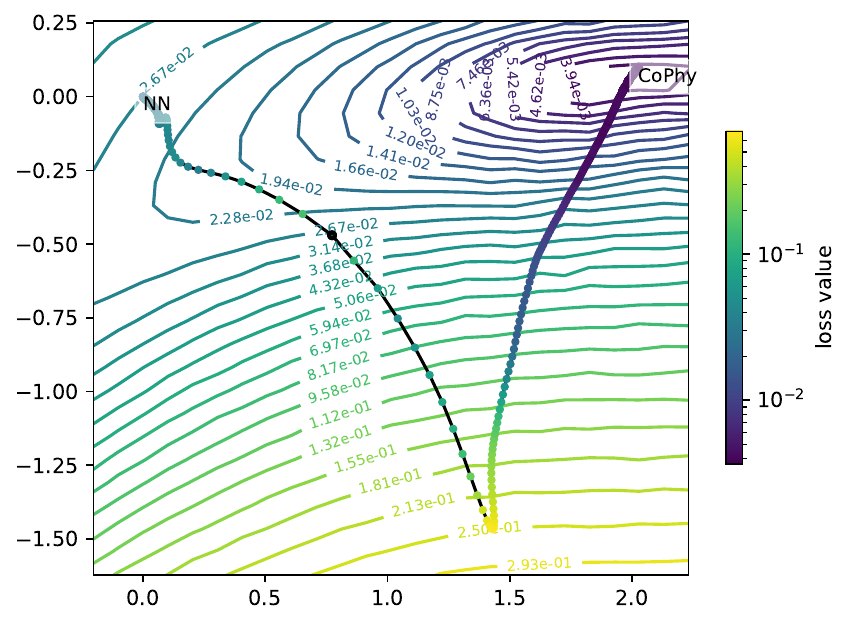}
                \caption{PCA - $\text{Test-MSE}$ }
                \label{subfig:pcatest}
              \end{subfigure}
            \begin{subfigure}[b]{0.3\textwidth}
                \includegraphics[width=\textwidth]{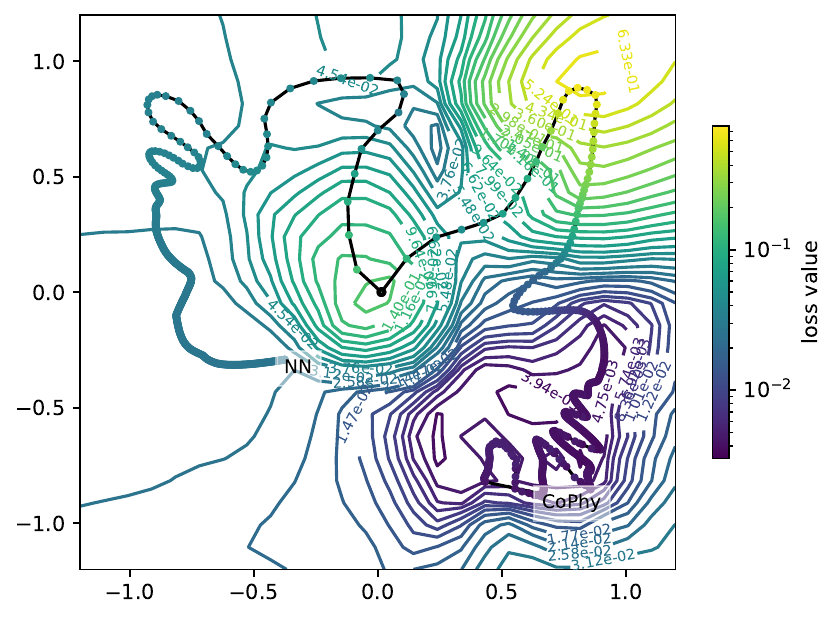}
                \caption{\proposedautencoder{} - $\text{Test-MSE}$ }
                \label{subfig:minetest}
              \end{subfigure}

         
              \begin{subfigure}[b]{0.3\textwidth}
                \includegraphics[width=\textwidth]{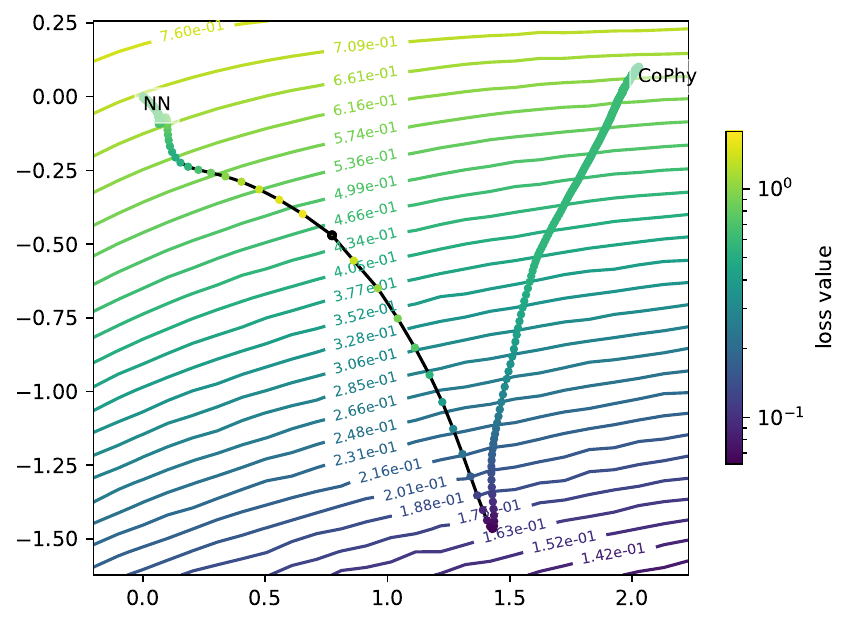}
                \caption{PCA - $\eloss{}$ }
                \label{fig:elossPCA}
              \end{subfigure}
              \begin{subfigure}[b]{0.3\textwidth}
                \includegraphics[width=\textwidth]{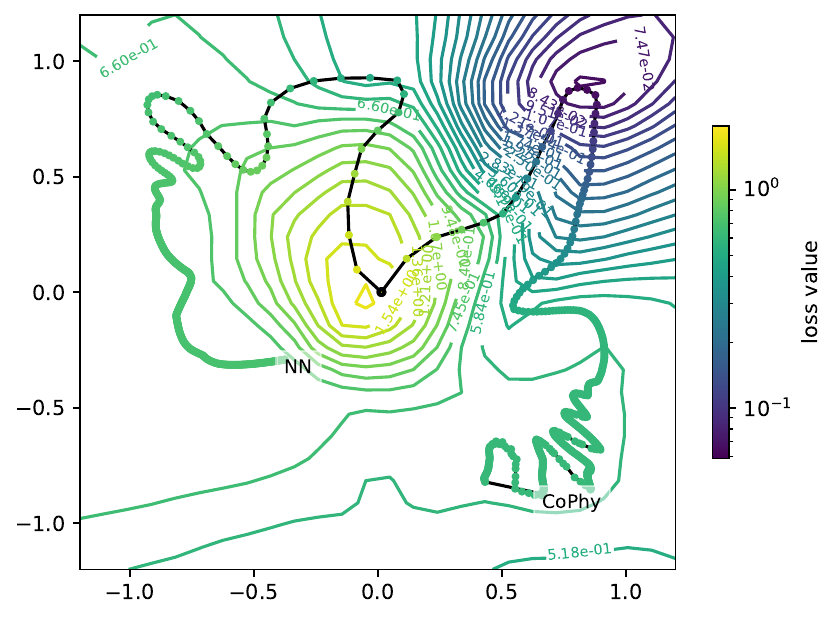}
                \caption{\proposedautencoder{} - $\eloss{}$  }
                \label{fig:elossmine}
              \end{subfigure}
         
              
              \caption{The loss landscapes of \cophy{} and \nn{} for different loss terms using PCA and \proposedautencoder{}. Comparing the two, it is clear that \proposedautencoder{} tells a more accurate and insightful story.}
              \label{fig:NNcophylandscape}
            \end{figure}

          We here show \proposedautencoder{}'s usefulness for comparing different models by plotting the trajectories of \cophy{} and its baseline \nn{} in \cref{fig:NNcophylandscape}. Both trajectories start from the same model initialization (marked with a thick border). The top two sub-figures use PCA to visualize $\text{Test-MSE}$ and the spectrum loss, \eloss{}, which is one of the two physics losses used to train the model. 
          The bottom two sub-figures show the corresponding loss landscapes for \proposedautencoder{}. We notice that 
          PCA utterly fails at capturing \nn{}'s critical points. In \cref{fig:elossPCA}), 
          PCA not only misses \nn{}'s minima, but also shows inconsistency in its loss values (i.e., the model colors indicate that it is descending, while the contours indicate the opposite). This is in contrast to \proposedautencoder{}, where \nn{}'s and \cophy{}'s minima is distinctly visualized with high consistency in terms of loss values. 
          Another inconsistency is found by looking at \nn{}'s trajectory in \cref{subfig:pcatest}. Here, PCA shows that it crosses the $2.28\mathrm{e}{-2}$ contour twice, implying that the trajectory descends and then ascends again. This, however, is false as can be verified by looking at the trajectory model colors and how they become darker as the final model is approached, indicating that the loss monotonically decreases. This contradiction leads to confusion
          that is not present in \proposedautencoder{}'s case in \cref{subfig:minetest}, making the latter more useful. 
          
                     
           \subsection{Applying \proposedautencoder{} on PINNs} 

            We here explore \proposedautencoder{}'s usefulness in studying Physics-Informed Neural Networks (PINNs), which have been widely and successfully used by many researchers in recent years to solve partial differential equations (PDEs), which appear in many real-world engineering and scientific applications. Moreover, PINNs present themselves as a convenient test-bed for our proposed landscape visualization method, thanks to 
            the significant effect of varying optimization hyper-parameters on PINN performance.

            For training PINNs, the loss terms that play a role  
            are the residual loss $\lres$, the initial condition loss $\lic$, and the boundary condition loss $\lbc$. The total loss being optimized is:
            \begin{equation}
                \text{L}_{total} = \cres \times \lres + \cic \times \lic + \cbc\times  \lbc.
            \end{equation}
            See \cref{app:pdes} for a detailed literature review on PINNs.

        \subsubsection{Demonstrating \proposedautencoder{}'s Flexibility With Different Constraints.} \label{ss:constraints}
        As discussed in \cref{se:constaints}, one of \proposedautencoder{}'s advantages is its ability to warp the learned manifold to satisfy certain constraints. To demonstrate this, we use PINN for the Convection equation as a target application. We set $\beta=30$, a high value, making the PDE harder to solve 
        and the loss landscape more complex and interesting to visualize (see \cref{app:pdes} for details).

\begin{figure*}[htb]
              \centering
              \begin{subfigure}[b]{0.3\textwidth}
                \includegraphics[width=\textwidth]{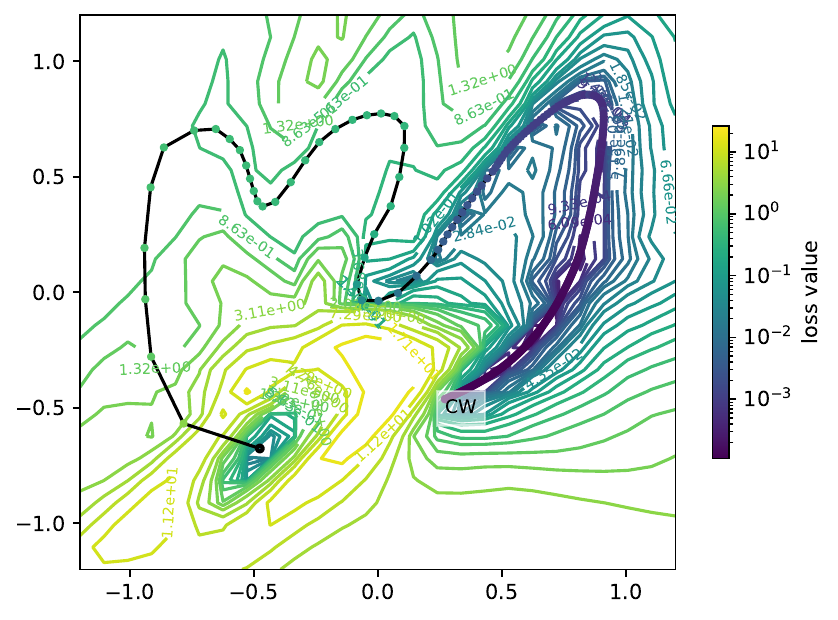}
                \caption{vanilla}
                \label{fig:constraintsvanilla}
              \end{subfigure}
              \hfill
            \begin{subfigure}[b]{0.3\textwidth}
                \includegraphics[width=\textwidth]{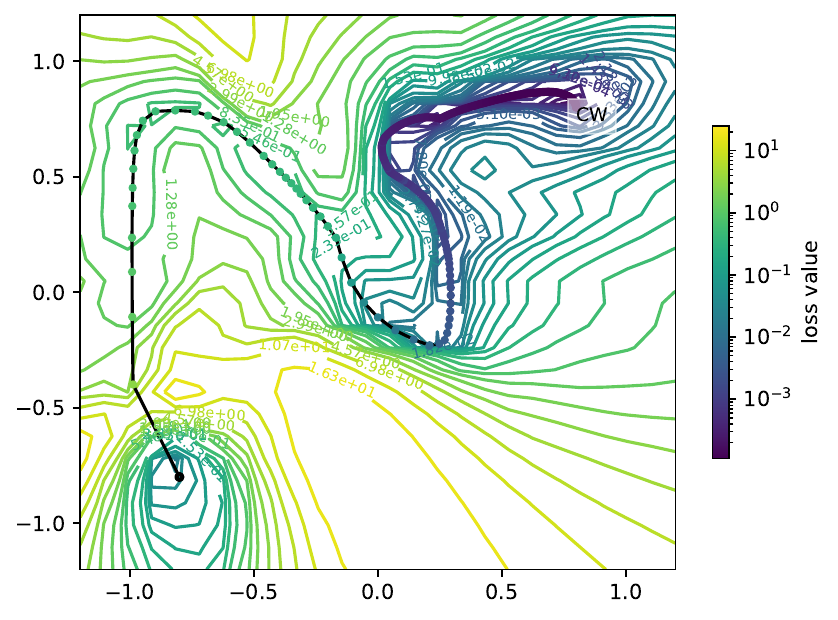}
                \caption{with ${\lanch}_1$}
                \label{fig:constraintspolar}
              \end{subfigure}
              \hfill
              \begin{subfigure}[b]{0.3\textwidth}
                \includegraphics[width=\textwidth]{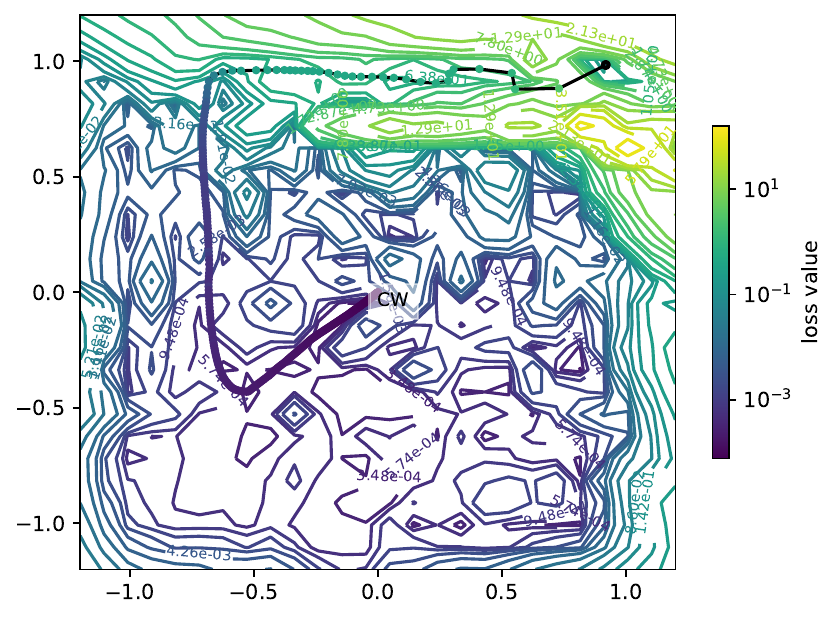}
                \caption{with ${\lanch}_2$ and $\lgrid$ }
                \label{fig:constraintgrid}
              \end{subfigure}

              \caption{A series of \proposedautencoder{} landscape visualizations of $\lres$ with different constraints. Notice the versatility of the proposed method and its ability to learn the desired manifold by employing appropriate constaints.}
              \label{fig:constraints}
        \end{figure*}

        \cref{fig:constraints} shows a progression of \proposedautencoder{} models visualizing $\ltest$  (i.e., the prediction error at test domain points) of the same PINN model. However, these \proposedautencoder{} models are trained with different constraints. First, \cref{fig:constraintsvanilla} shows the loss landscape with no constraints. Subsequent sub-figures show the different manifolds obtained by varying the \proposedautencoder{}'s training constraints.
        \cref{fig:constraintspolar} uses ${\lanch}_1$. As a result, the trajectory stretches almost perfectly across the grid between two opposite corners. 
        Alternatively, to show the effect of $\lgrid$, 
        we use a large $\lmax=8$ to impose high grid density at the vicinity of the trajectory models. As expected, \cref{fig:constraintgrid} shows a grid that zooms almost entirely onto
        the vicinity of the trajectory, highlighting more details of that area compared to the previous sub-figures.

\begin{figure}[htb]
          \centering
        \begin{subfigure}[b]{0.23\textwidth}
            \includegraphics[width=\textwidth]{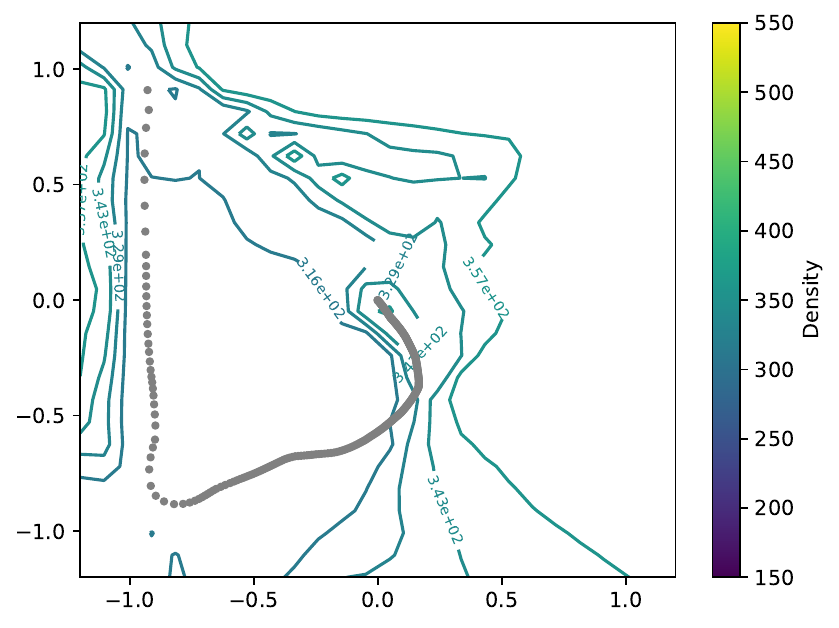}
            \caption{$L=2$}
          \end{subfigure}
          \hfill
          \begin{subfigure}[b]{0.23\textwidth}
            \includegraphics[width=\textwidth]{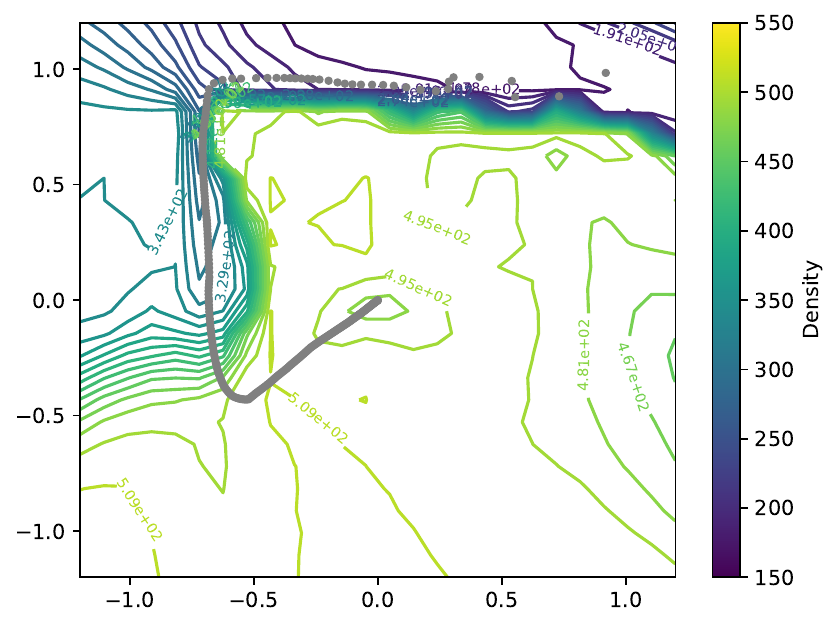}
            \caption{$L=8$}
          \end{subfigure}

          \caption{A comparison of two \proposedautencoder{} \textit{density} landscape visualizations as a result of using $\lgrid$ with two different $\lmax$ values. As can be seen, through this constraint, \proposedautencoder{} grants its user greater flexibility to decide the appropriate zooming factor.}
          \label{fig:Lgrid}
        \end{figure}

        We further study the effect of $\lgrid$ by varying $\lmax$, the hyper-parameter that correlates with the grid density near trajectory models. \cref{fig:Lgrid} shows \underline{the \textit{density} landscape} (i.e., the color-coding indicates grid density and not loss values) for two different values of $\lmax$. 
        We use a $\text{CKA}$-similarity-based density that is defined as:


        \begin{equation}
            \rho_{m \in \gridpoints} = \sum_{m^\prime \in \gridpoints \setminus \{m\}} \text{CKA}(m^\prime, m),
        \end{equation}
        where $\text{CKA}(m^\prime, m)$ is the $\text{CKA}$ similarity measure between two neural networks as defined in \cite{kornblith2019similarity}.
        As can be seen in \cref{fig:Lgrid}, the density of the grid especially near trajectory models increases with $\lmax$. 
        This shows that $\lgrid$ can be a vital tool for engineering the manifold visualization through the appropriate choice of $\lmax$.

         \subsubsection{Using \proposedautencoder{} To Study PINNs' Training Pathologies.} \label{ss:NTK}
\begin{figure*}[htb]
            \centering
            
            \begin{subfigure}[b]{0.3\textwidth}
                \centering
                \includegraphics[width=\textwidth]{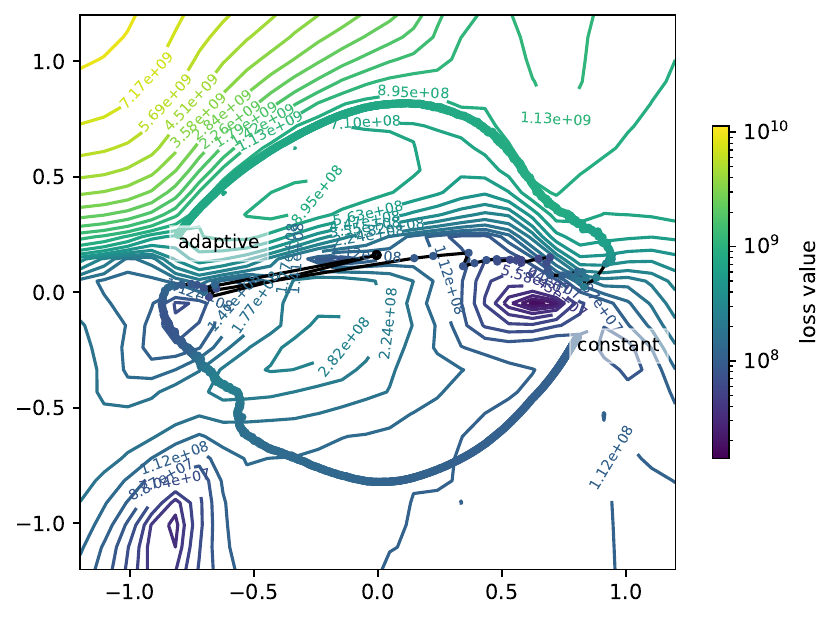}
                \caption{residual loss ($\lres$)}
            \end{subfigure}
            \hfill
            \begin{subfigure}[b]{0.3\textwidth}
                \centering
                \includegraphics[width=\textwidth]{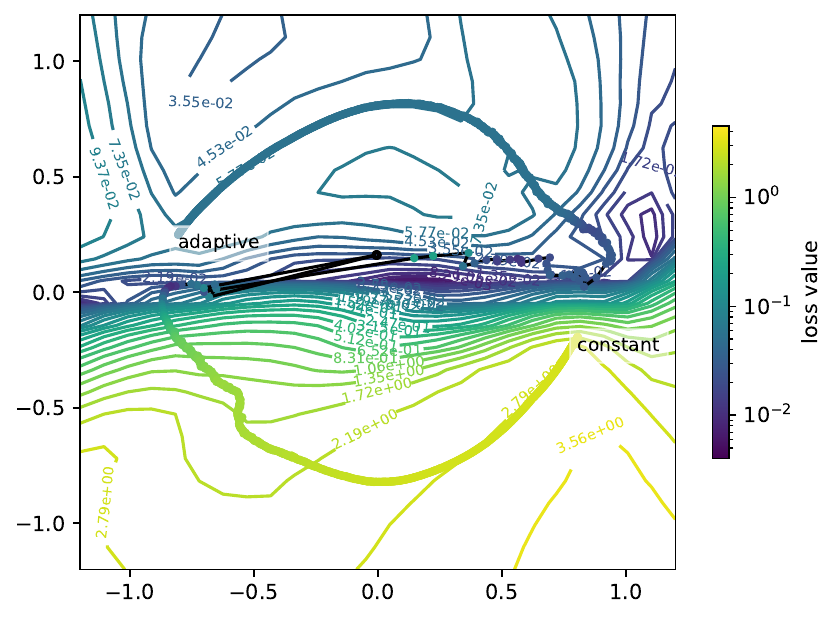}
                \caption{boundary condition loss ($\lbc$)}
            \end{subfigure}
            \hfill
            \begin{subfigure}[b]{0.3\textwidth}
                \centering
                \includegraphics[width=\textwidth]{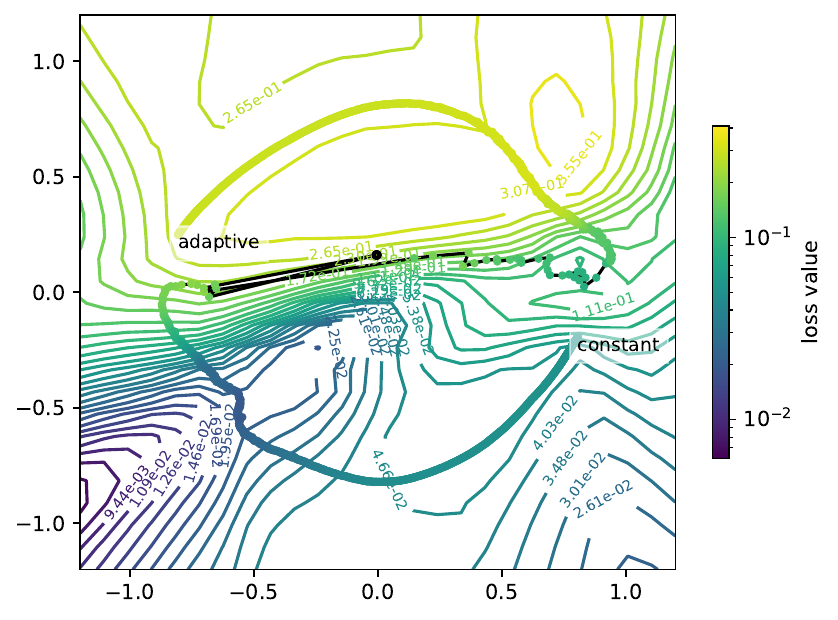}
                \caption{initial condition loss ($\lic$)}
            \end{subfigure}
            
            \caption{A comparison of the two optimization approaches studied by \cite{wang2020and} using \proposedautencoder{}. This visualization verifies the authors' claims and shows that an NTK-based adaptive approach emphasizes on a better optimization of the boundary condition loss and reaches flatter minima across all losses.}
            \label{fig:PINNsNTK}
        \end{figure*}

         Next, we use \proposedautencoder{} to visually verify \cite{wang2020and}'s findings on PINNs through the lens of Neural Tangent Kernel (NTK). In their work, the authors hypothesize that PINN training pathologies result from a discrepancy in the convergence rate of the different loss components. More precisely, they show that the PDE's residual loss ($\lres$) converges faster than the boundary condition loss ($\lbc$), leading to a sub-optimal model. To verify this claim, we train a \proposedautencoder{} to visualize the loss landscape for the two different optimization approaches considered in \cite{wang2020and}. Namely, an approach that trains with constant loss weighting, and another with NTK-based adaptive loss weighting. An ${\lanch}_3$ pinning constraint is imposed to place the models over the perimeter of a circle, making it easy to compare the two approaches. As can be seen in \cref{fig:PINNsNTK}, the authors' claim is easily verifiable using our proposed visualization method. First, in line with the authors' hypothesis, while the terminal $\lres$ and $\lic$ are relatively worse for the NTK-based approach, its terminal $\lbc$ is much better, indicating that an optimal $\lbc$ is essential for obtaining a well-trained PINN. Another observation we make in accordance with their paper is that the NTK-based approach reaches flat minima for all losses at a somewhat similar cadence. This is in contrast to the baseline model where the different losses converge at variable rates, traversing loss landscapes that are less flat and of variable slopes.

           \subsubsection{Using \proposedautencoder{} To Study PINNs' Failure Modes.} \label{se:mahoney}

           Inspired by \cite{krishnapriyan2021characterizing}'s work on understanding the effect of the PDE's complexity and PINN's regularization 
          on the loss landscape and optimization outcome, we use \proposedautencoder{} to study PINN performance on the convection equation. Namely, we want to validate whether a higher $\beta$ parameter in a convection PDE or an increase in PINN regularization (i.e., an increase in the value of $\cres$) leads to a more complex loss landscape that is harder to optimize. As such, we run an experiment where we train PINNs with varying $\beta$ values, and then compare their loss landscapes using \proposedautencoder{}. A similar experiment on $\cres$ can be found in \cref{app:cres}. 

\begin{figure*}[htb]
            \centering
            
            \begin{subfigure}[b]{0.3\textwidth}
                \centering
                \includegraphics[width=\textwidth]{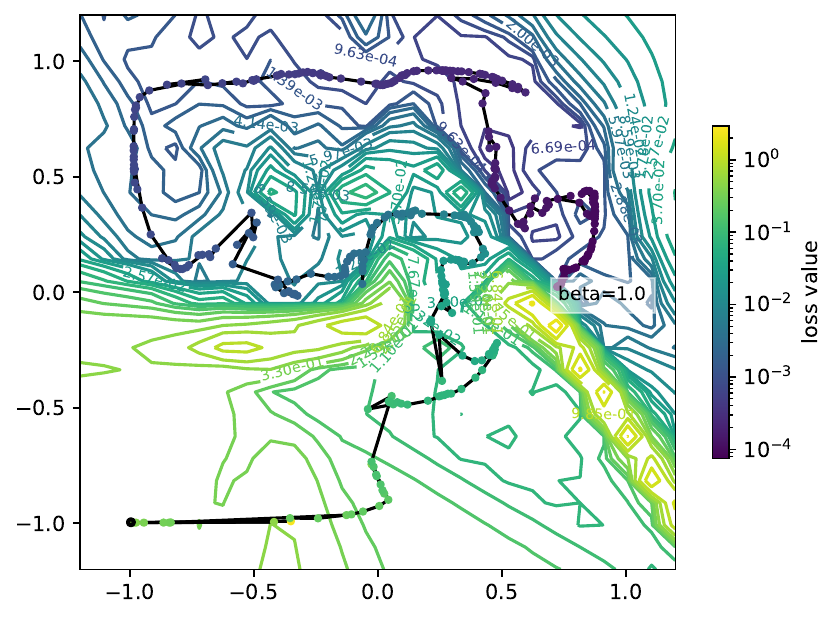}
                \caption{$\beta=1$}
            \end{subfigure}
            \hfill
            \begin{subfigure}[b]{0.3\textwidth}
                \centering
                \includegraphics[width=\textwidth]{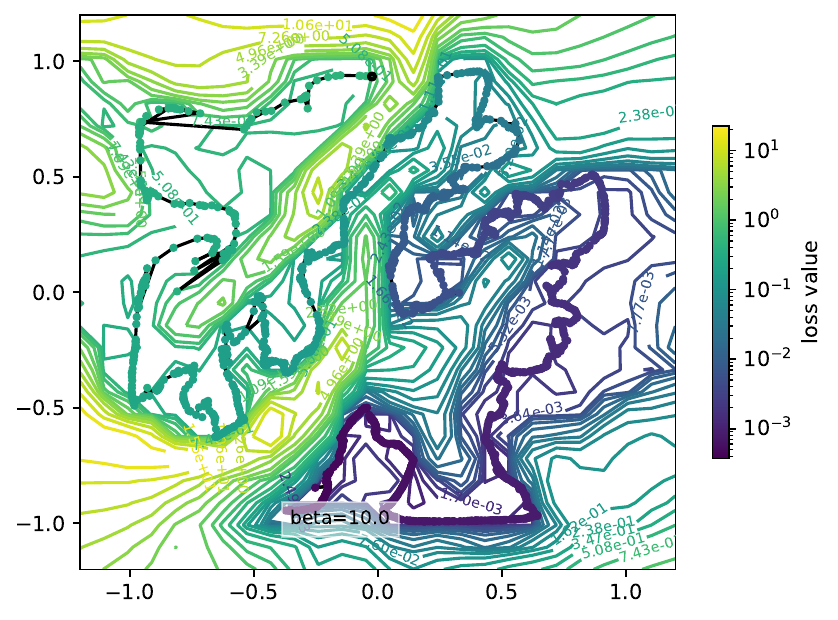}
                \caption{$\beta=10$}
            \end{subfigure}
            \hfill
            \begin{subfigure}[b]{0.3\textwidth}
                \centering
                \includegraphics[width=\textwidth]{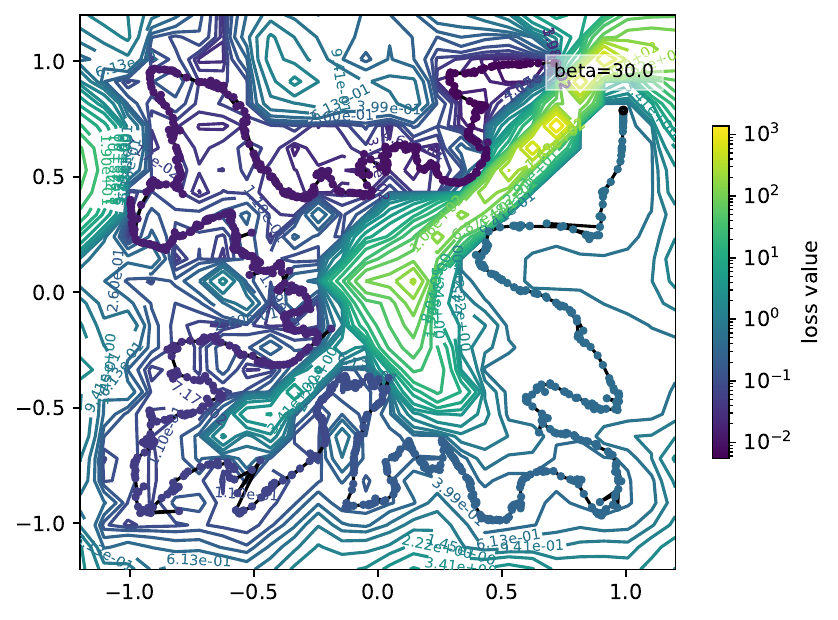}
                \caption{$\beta=30$}
            \end{subfigure}
            
            \caption{A comparison of PINNs solving convection PDEs of varying $\beta$s. \proposedautencoder{} verifies the claim of \cite{krishnapriyan2021characterizing} that increasing $\beta$ causes $L_{total}$'s landscape to become more non-convex and difficult to optimize.}
            \label{fig:FailsBeta}
        \end{figure*}

          Looking at \cref{fig:FailsBeta}, it is easy to verify that an increase in $\beta$ renders the loss landscape more non-convex and harder to optimize. This agrees with the authors' findings. However, it is worth noting that the loss landscape visualizations presented in \cite{krishnapriyan2021characterizing}  used linear methods. When \cref{fig:FailsBeta} is compared to its counterpart (i.e.,  Figure 3 in \cite{krishnapriyan2021characterizing}), it is evident that their loss landscape visualizations are
          less intuitive and hard to visually interpret without the authors' commentary, indicating that our proposed method is more effective.

           \subsubsection{Investigating Different Loss Balancing Techniques.} \label{sss:MTL}
           Balancing different loss terms in frameworks such as multi-task learning (MTL) or \SGML{} can be a daunting task \cite{wang2020and,doi:10.1137/20M1318043,elhamod2022cophy}. Thus, the ability to compare different loss balancing techniques is an important research effort. Generally, different loss balancing techniques are compared based on the final model's accuracy. 
           However, this metric 
           does not provide a fundamental understanding of the optimization process 
           and whether a loss balancing algorithm is more suitable for one specific task than another. 

           Here, we consider six different loss balancing methods that are commonly used in the literature; namely Equal Weights (\ew{}), Constant Weights (\cw{}), Dynamic Weight Averaging (\dwa{}), Learning Rate Annealing \cite{doi:10.1137/20M1318043} (\lranneal{}), Gradient Normalization \cite{pmlr-v80-chen18a} (\gradnorm{}), and Random Loss Weighting \cite{lin2022reasonable}  (\rlw{}). 
           \cref{app:lossbalancing} gives a detailed account of these algorithms.

           To compare these algorithms, starting from the same model initialization, we train multiple PINNs with the listed algorithms to solve the Convection problem with $\beta=10$. 

\begin{figure}[H]
            \centering



              \begin{subfigure}[b]{0.3\textwidth}
                \includegraphics[width=\textwidth]{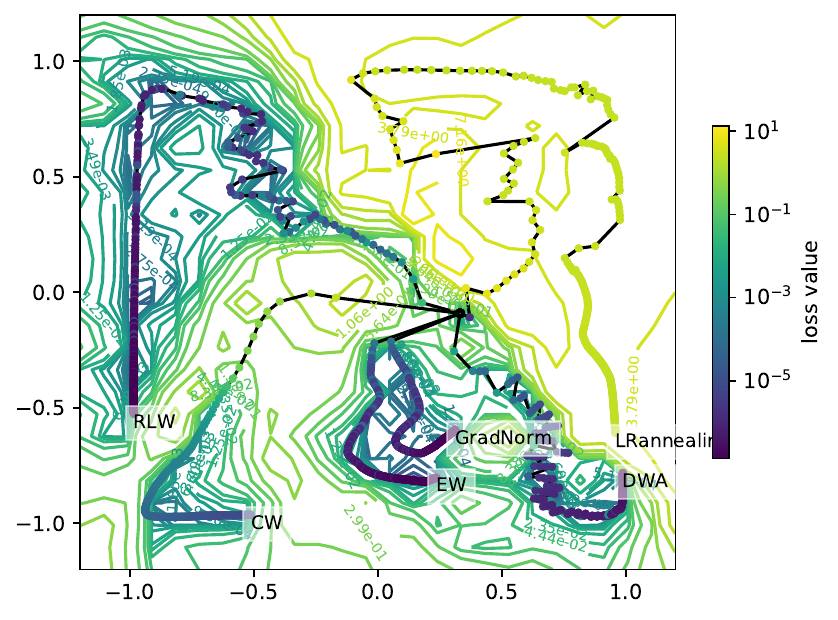}
                \caption{$\lres$}
                \label{subfig:residualAE}
              \end{subfigure}
              \begin{subfigure}[b]{0.3\textwidth}
                \includegraphics[width=\textwidth]{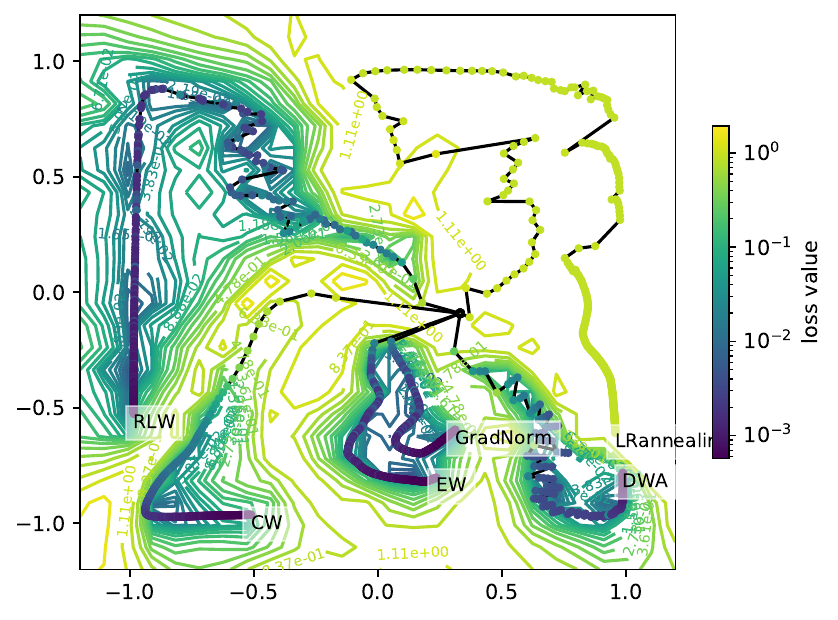}
                \caption{$\ltest$}
                            \label{subfig:testAE}
              \end{subfigure}

              \caption{\proposedautencoder{}'s loss landscapes of PINN models with different loss balancing methods intuitively and visually provide valuable insights that would otherwise be hard to extract.}
              \label{fig:dwa}
            \end{figure}

        In \cref{fig:dwa}, we inspect the landscapes of two loss terms: $\lres$ and $\ltest$. 
        Using \proposedautencoder{}, 
        the different trajectories and minima can be easily found and compared. In particular, it is clear that while \gradnorm{} and \ew{} take slightly different trajectories, they converge to the same minima. On the other hand, \dwa{} converges to the same basin as \gradnorm{} and \ew{} w.r.t $\lres$, but not the same minima.  
        Additionally, looking at \cref{subfig:testAE}, it is clear that \lranneal{}'s trajectory is nowhere near a minima w.r.t $\ltest$. This is surprising since \lranneal{} has shown success at solving the Helmholtz equation and the Klein Gordon equation as demonstrated in \cite{doi:10.1137/20M1318043}, implying that not all PINN tasks benefit from this method. 
        The valuable insight that was visually, easily, and intuitively inferred from \proposedautencoder{} in \cref{fig:dwa} would not have been possible with a baseline method such as PCA due to its linear scale and planar manifold. \cref{app:MTLPCA} shows the equivalent results using PCA to verify this claim.



            \section{Conclusions and Future Work} \label{se:conclusion}
                In this paper, we have shown that our proposed auto-encoder-based method, \proposedautencoder{}, is capable of learning non-linear manifolds in the input model parameter space and mapping such manifolds onto a 2-D grid for loss landscape visualization. Additionally, we have demonstrated that \proposedautencoder{} surpasses many other linear and non-linear landscape visualization approaches in terms of representation accuracy and malleability to learning manifolds with user-defined properties. Finally, we used two applications, \cophy{} \cite{elhamod2022cophy} and PINNs \cite{raissi2017physics1}, to derive insightful findings, including the importance of certain hyper-parameters for neural network training and the efficacy of different deep learning frameworks such as Multi-Task Learning (MTL) and Knowledge-Guided Machine Learning (\SGML). 
                Future work could explore the full potential of \proposedautencoder{} by using it to study other deep learning phenomena, such as the relationship between landscape sharpness and generalizability \cite{pmlr-v137-huang20a}.
                
                One of \proposedautencoder{}'s limitations is the lack of scalability for large input parameter spaces. For example, the PINNs studied in this paper have an input dimensionality of the order of $10,000$ parameters, which is relatively small. In contrast, computer vision applications generally use much larger models (e.g., a VQ-GAN \cite{esser2021taming} has approximately $88$ million parameters).
                Such a large input space would be prohibitively difficult for \proposedautencoder{} to encode directly. A future research direction could investigate appropriate ways to encode and visualize such large models effectively. 

\bibliography{sample-base}
            
\clearpage

\appendix

    \section{Hyper-parameter Selection and Additional Implementation Details} \label{app:hyper}
            To guarantee that an input model $m \in \manifold$ is encoded into $z \in [-1, +1]\times [-1, +1]$, a \textit{tanh} activation function is appended to the output of the encoder $\encoderlandscape$. Also, $m$ is normalized for an easier training of $\proposedautencodermodel$.

            With regards to location anchoring constraints $\lanch$, we here define the three different examples mentioned in \cref{p:anchoring} more rigorously: 

            \begin{itemize}[itemsep=0em]
            \item \textit{Polar pinning (${\lanch}_1$) }: This can be formally defined by setting $\trajectorymodels^\prime = \{ m_{0}, m_{\left| \trajectorymodels \right|} \}$ (i.e., the set that contains the first and last models in the trajectory), and $\mathcal{A} = \{ (-r, -r), (r, r)\}$, with an $r=0.8$.
            \item \textit{Center pinning (${\lanch}_2$) }: It is formally defined with $\trajectorymodels^\prime = \{ m_{\left| \trajectorymodels \right|} \}$ and $\mathcal{A} = \{(0, 0)\}$.
            \item \textit{Circle pinning (${\lanch}_3$) }: It is formally defined with $\trajectorymodels^\prime = \{ m_{\left| \trajectorymodels \right|} \}$ and $\mathcal{A} = \{(r \cdot \text{sin}(\frac{2 \pi k}{n}) , r \cdot \text{cos}(\frac{2 \pi k}{n}))\ \text{where } k \in \{0, 1, ...,  n-1\}  \}$
            \end{itemize}


            It is worth mentioning that we use another constraint, $\ltraj$, in some of our experiment to help space the trajectory models equally so that they are spread out across the grid. The loss term for this constraint penalizes for the differences in step sizes in the latent grid space between consecutive trajectory models.
            
            
            

            \cref{tbl:hyper} shows the hyper-parameters used to train \proposedautencoder{} for both \cophy{} and PINNs. 

            \begin{table*}[htb]
            \begin{tabular}{|p{3cm}|l|l|l|l|l|}
            \hline
                           & encoder hidden layer sizes & batch size & epochs & lr   & notes                                                                                                                                   \\ \hline
            \cophy{}. \cref{ss:cophy_landscape2}         & $3141, 270, 23$               & $32$         & $40,000$  & $10^{-4}$ &                                                                                                                                         \\ \hdashline
            PINNs - "Investigating Different Loss Balancing Techniques". \cref{sss:MTL}  & $991, 125, 15$                 & $32$         & $600,000$ & $5 \times 10^{-4}$ & $\crec{} = 10^4$                                                                                                      \\ \hdashline
            PINNs - "Using \proposedautencoder{} To Study PINNs' Training Pathologies". \cref{ss:NTK}  & $1000, 500, 100$              & $32$         & $100,000$  & $10^{-5}$ & $\canch{}_3 = 10^4$                                              \\ \hdashline
            PINNs - "Using \proposedautencoder{} To Study PINNs' Failure Modes". \cref{se:mahoney} & $991, 125, 15$                 & $32$         & $600,000$ & $5 \times 10^{-4}$ & $\ctraj{} = 1$                                                                                                \\ \hdashline
            PINNs - "Demonstrating \proposedautencoder{}'s Flexibility With Different Constraints".  \cref{ss:constraints}  & $991, 125, 15$                 & $32$         & $80,000$  & $10^{-4}$ & \begin{tabular}[c]{l} $\canch{}_1 = 10^2$ \\ \\ $\canch{}_2 = 10^2$, $\cgrid{} = 1$ \end{tabular} \\ \hline
            \end{tabular} 

            \caption{A table of the hyper-parameter values used to train \proposedautencoder{} for each experiment.}
            \label{tbl:hyper}
            \end{table*}

            Finally, the experiments in this paper were run on Nvidia DGX A100 GPUs. Each experiment needed a single GPU and 8 CPU cores.

  \section{On the ``Correctness” of Loss Landscape Visualization Methods} \label{app:correctness}
            
           Loss landscape visualization is a subjective tool that provides a qualitative and holistic description of the landscape, rather than a quantitative one. As such, the notion of ``correctness" is not the best vantage point from which this tool can be appreciated. One analogy that clarifies this point is map projections. Earth map projections represent the 3-D Earth surface on a 2-D plane. Different projection methods have different advantages and limitations; each method makes certain features or characteristics of the Earth surface more or less prominent. Thus, there are many valid projections, each useful for different purposes. One popular method is the Mercator projection \cite{snyder1997flattening}, which is useful for navigation because it preserves angles and directions. However, it distorts the size and shape of objects near the poles. Another method is the Robinson projection \cite{snyder1990robinson}, which balances distortions of size and shape, making it a good all-purpose projection. 
           A third method is the Winkel tripel projection \cite{snyder1997flattening} which 
           accurately shows the relative sizes and shapes of landmasses, but still distorts the shapes of some landmasses and oceans. 
           
           Similarly, a non-linear loss landscape visualization method will produce a non-linear manifold that gets distorted when visualized on a 2-D planar grid. . Thus, while the user should be aware of these distortions and understand the visualization accordingly, the method is still valid and useful for understanding the properties of the loss surface. And while there are infinitely many manifolds that could contain a set of points in the parameter space (i.e., models), it is important to select the manifold(s) that exhibits the desired user-defined criteria, such as the scaling factors at different parts of the manifold.

    \section{A Brief Description of \cophy{}} \label{app:cophy}

\cophy{} \cite{elhamod2022cophy}, short for Competing Physics Physics-Guided Neural Networks, is a model uniquely tailored to solve eigenvalue problems, which are prevalent in scientific domains such as quantum mechanics and electromagnetic propagation. In addition to the data-drive $\text{Train-Loss}$, two physics-guided (PG) loss terms are used: \text{\nsloss{}} and \text{\eloss{}}.\text{\nsloss{}} is designed to enforce the eigen-equation, and its minimization can lead to multiple solutions that satisfy the physical constraints of the problem. However, these multiple solutions may correspond to different energy levels. Generally, only one solution is desired. This is where S-Loss comes into play. \text{\eloss{}} guides the network towards the specific energy level of interest by minimizing the difference between the predicted and target eigenvalues

The overall learning objective of \cophy{} is:
\begin{equation}
    E(t) = \text{Train-Loss} + \lambda_C(t) ~ \text{\nsloss{}} + \lambda_S(t)  ~\text{\eloss{}}
\end{equation}

The methodology of \cophy{} involves adaptively tuning the coefficients of these loss functions by annealing \( \lambda_S \) so as to steer the model towards the correct energy level in the initial stages of training. Conversely, \( \lambda_C \) is cold started, allowing the physical constraints to be gradually enforced and ensuring that the solution adheres to the underlying physics.

Using two applications – predicting the ground-state wave function of an Ising chain model in quantum mechanics, and modeling electromagnetic wave propagation in periodically stratified layer stacks – the authors demonstrate the effectiveness and extrapolative power of \cophy

    \section{Loss Landscapes Error Plots For \cophy{}} \label{app:errors_cophy}

\begin{figure}[htbp]
              \centering
              \begin{subfigure}[b]{0.3\textwidth}
                \includegraphics[width=\textwidth]{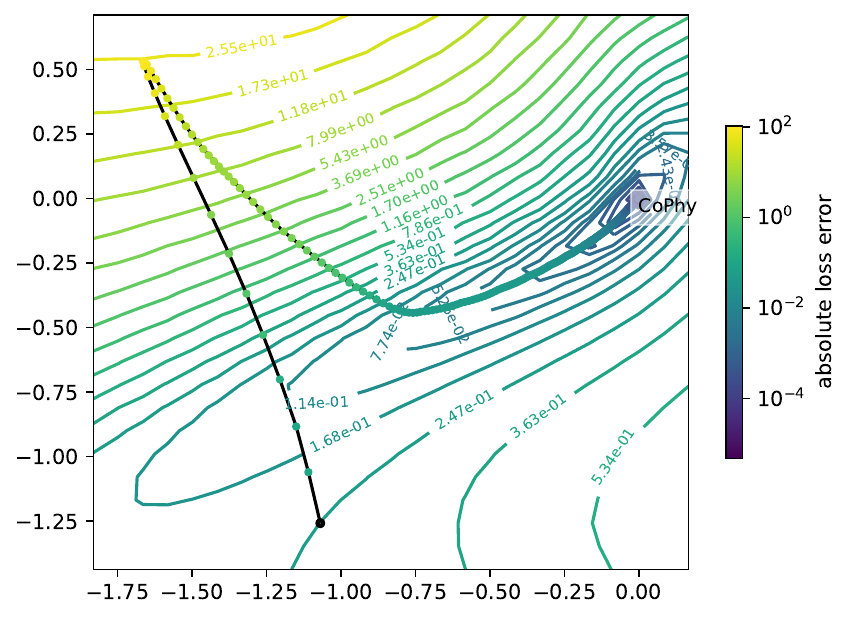}
                \caption{PCA}
                \label{fig:PCAvsAE2_PCAlosses}
              \end{subfigure}
              \begin{subfigure}[b]{0.3\textwidth}
                \includegraphics[width=\textwidth]{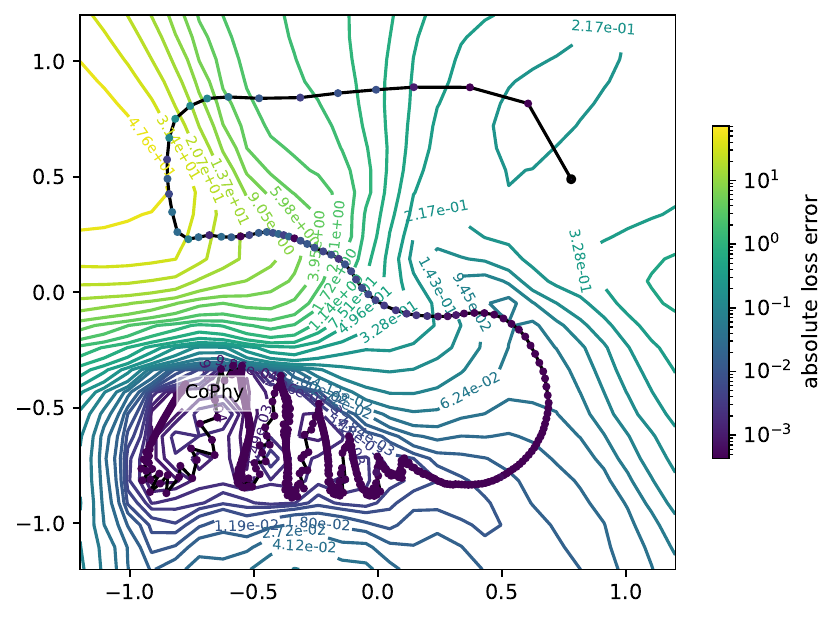}
                \caption{\proposedautencoder{}}
                \label{fig:PCAvsAE2_losses}
              \end{subfigure}
              \begin{subfigure}[b]{0.3\textwidth}
                \includegraphics[width=\textwidth]{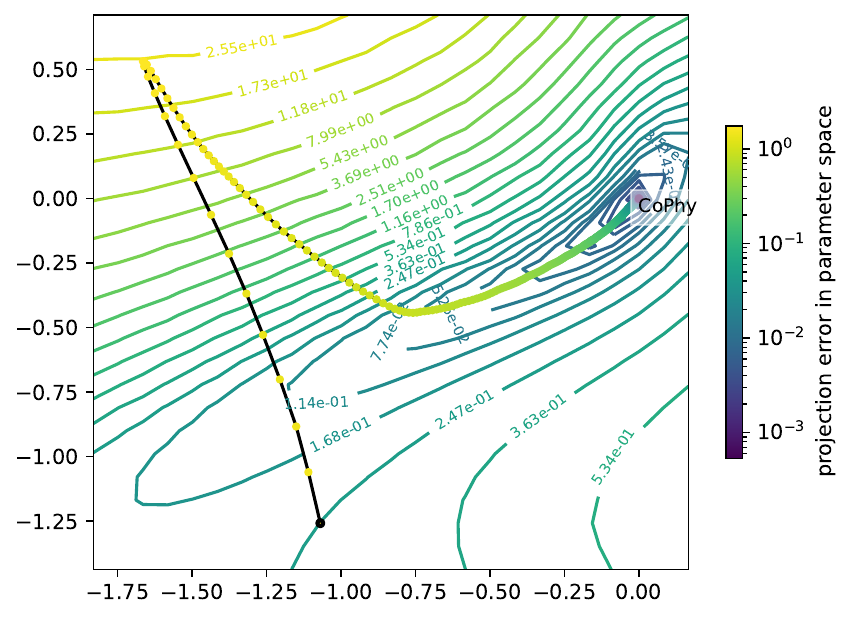}
                \caption{PCA}
                \label{fig:PCAvsAE2_PCAdistances}
              \end{subfigure}
              \begin{subfigure}[b]{0.3\textwidth}
                \includegraphics[width=\textwidth]{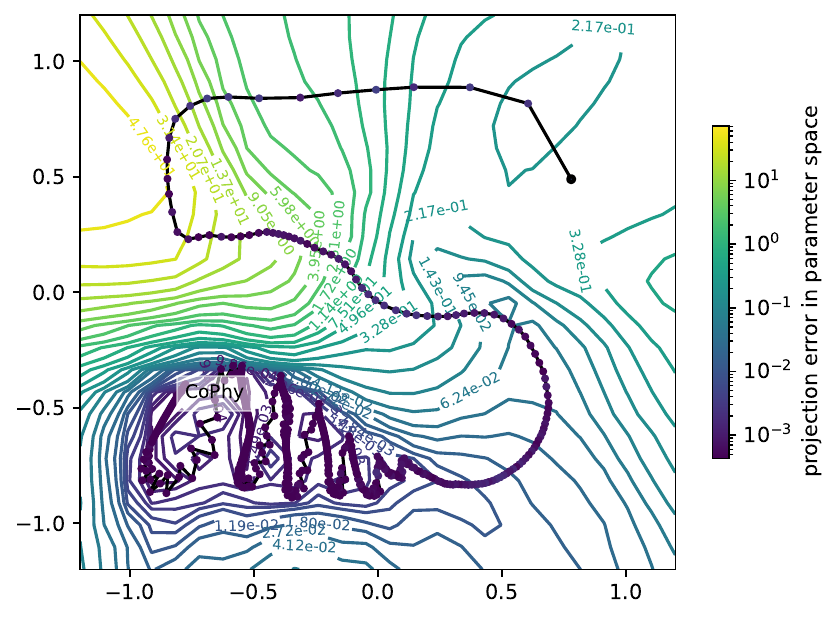}
                \caption{\proposedautencoder{}}
                \label{fig:PCAvsAE2_distances}
              \end{subfigure}

              \caption{A comparison of PCA and \proposedautencoder{} in terms of the error in \cophy{}'s total physics loss values (top two sub-figures) as well as projection errors (bottom two sub-figures). Note that the contour colors still refer to the loss values, similar to \cref{fig:PCAvsAE}. \proposedautencoder{} has lower error levels than PCA.}
              \label{fig:PCAvsAE2}
            \end{figure}

\begin{figure}
              \centering
              \begin{subfigure}[b]{0.3\textwidth}
                \includegraphics[width=\textwidth]{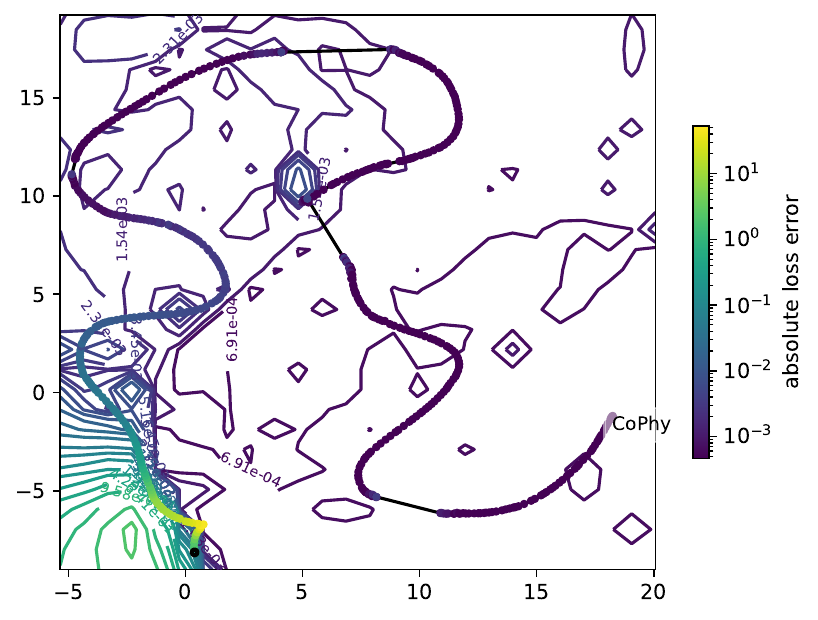}
                \caption{UMAP}
              \end{subfigure}
              \begin{subfigure}[b]{0.3\textwidth}
                \includegraphics[width=\textwidth]{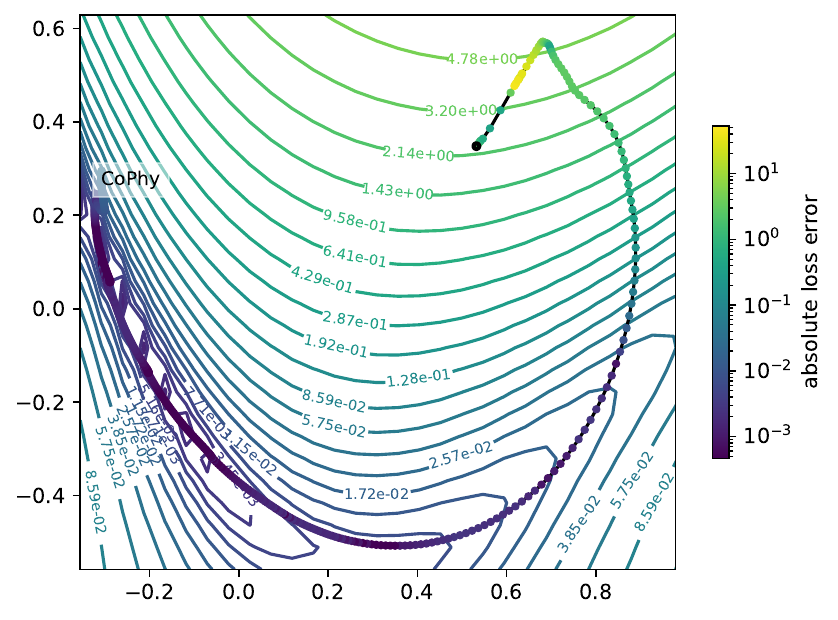}
                \caption{Kernel-PCA}
              \end{subfigure}
              \begin{subfigure}[b]{0.3\textwidth}
                \includegraphics[width=\textwidth]{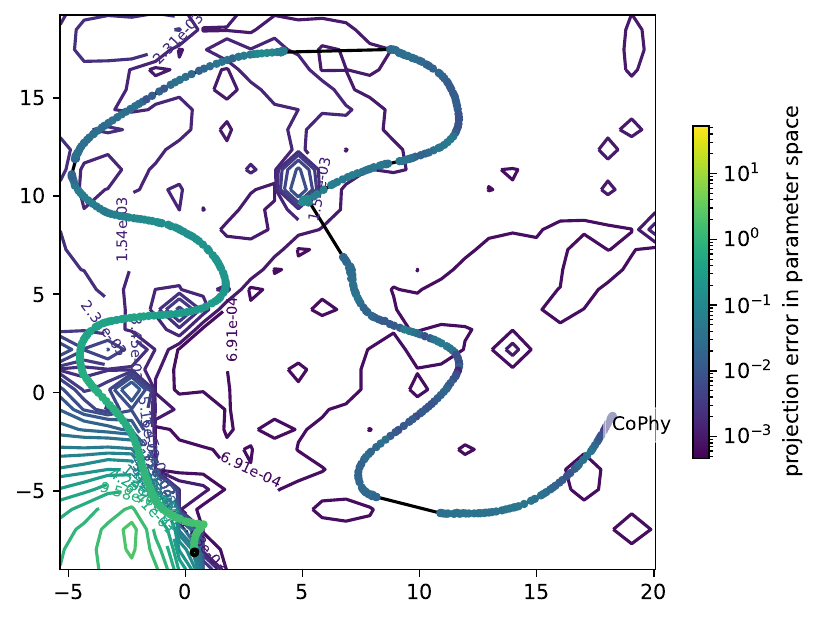}
                \caption{UMAP}
              \end{subfigure}
              \begin{subfigure}[b]{0.3\textwidth}
                \includegraphics[width=\textwidth]{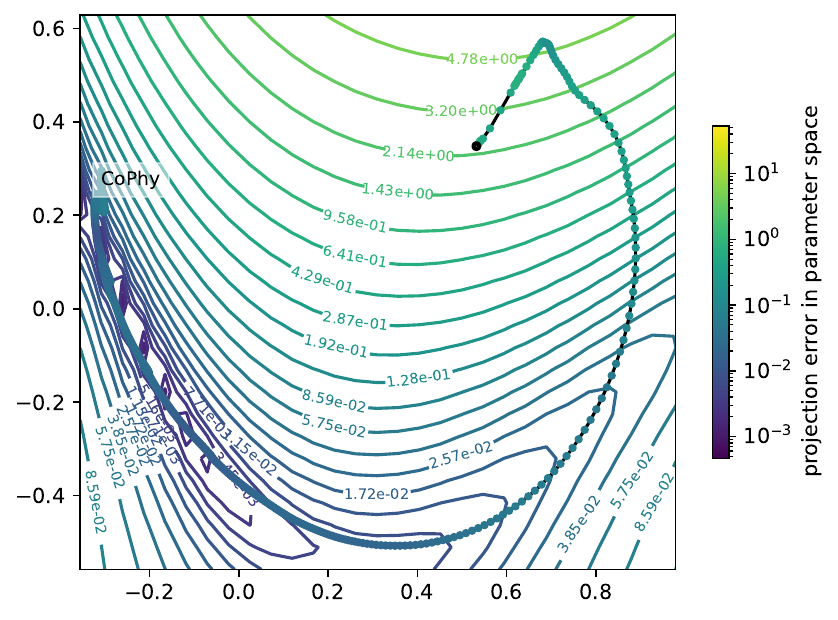}
                \caption{Kernel-PCA}
                \label{fig:subfig2}
              \end{subfigure}

              \caption{A comparison of UMAP and Kernel-PCA in terms of the error in \cophy{}'s total physics loss errors (top row) and projection errors (bottom row). Compared to \cref{fig:PCAvsAE2}, neither UMAP nor Kernel-PCA look favorable against \proposedautencoder{}. Note that the contour colors still refer to the loss values, similar to \cref{fig:PCAvsAE}. 
              }
              \label{fig:UMAPKernelPCA_app}
            \end{figure}

    Following up on \cref{fig:PCAvsAE}, and to visually illustrate the results of \cref{tab:quant_losslandscape}, \cref{fig:PCAvsAE2,fig:UMAPKernelPCA_app} show the absolute loss error (i.e., the error between model loss and the loss at its projection on the learned manifold) in the first row, 
    and the distance between the models and the manifold \underline{in the parameter space} in the second row. In both cases, the ``hot" model colors of baseline methods, compared to \proposedautencoder{}, indicate that the resulting manifold or plane is not a good fit for the trajectory.

   \section{Solving Partial Differential Equations with Physics-Informed Neural Networks (PINNs)} \label{app:pdes}

        Solving partial differential equations (PDEs) is a fundamental problem in many areas of science and engineering. Traditional numerical methods, such as finite element and finite difference methods, require a discretization of the domain and the PDEs, which can lead to high-dimensional and computationally expensive systems. Physics-informed neural networks (PINNs) \cite{raissi2017physics2, raissi2019physics, raissi2017physics1, GAO2021110079} have emerged as an alternative approach to solving PDEs, leveraging the representational power of neural networks and the structure of the underlying physics to learn a solution without discretization.

        PINNs have shown great promise in solving various types of PDEs, including elliptic, parabolic, and hyperbolic problems. The main idea behind PINNs is to parameterize the solution of a PDE with a neural network and enforce the governing equations as constraints in the training process. This is achieved by incorporating the PDEs as a loss function that is minimized during training, along with a data-driven loss term that incorporates observed data. 

        PINNs' optimization objective comes in a number of different forms. Generally, however, there are three main losses that are present. 
        The residual loss, $\lres$, is the most important term and ensures that the neural network satisfies the PDE at every point in the domain. It is calculated as the mean squared difference between the residual of the PDE and the output of the neural network at each point. $\lic$ ensures that the neural network satisfies the PDE at the initial condition. It is calculated as the mean squared difference between the output of the neural network at the initial time and the true initial condition. $\lbc$ ensures that the neural network satisfies the PDE at the boundary conditions. It is calculated as the mean squared difference between the output of the neural network at each boundary point and the true boundary condition. Together, these three loss terms provide a comprehensive approach to training PINNs to solve PDEs. By balancing these terms, the neural network can learn the underlying physics of the problem and provide accurate solutions.

        PINNs have been shown to be effective in solving a wide range of partial differential equations (PDEs). For example, in fluid mechanics, PINNs have been used to solve problems related to incompressible Navier-Stokes equations \cite{JIN2021109951}. Similarly, PINNs have also been used to solve the Schrödinger equation in quantum mechanics \cite{li2022mix}.

        This paper has targeted a certain type of PDEs called the ``convection equation" to illustrate the usefulness and superiority of the proposed model. The convection problem is a type of partial differential equation that arises in fluid mechanics and heat transfer. It models the transport of a quantity (such as mass, energy, or momentum) by a moving fluid, which can induce a net flow in the direction of the transport. The convection equation is:
        
        \begin{equation}
            f = u_t - \beta u_x
        \end{equation}

        where the parameter $\beta$ is the convection coefficient which represents the tendency of the substance to move with the fluid flow. 

        While PINNs have been useful at solving PDEs, the application of PINNs is not without challenges \cite{wang2020and,doi:10.1137/20M1318043}. In terms of the convection problem, one major challenge is the presence of the convection term $\beta$ in the governing equations, which is highly nonlinear and can result in optimization difficulties, especially for higher values. Another challenge is the presence of multiple scales in convection problems, which can lead to numerical instability and slow convergence.

    \section{The Pathological Effect of Increasing Regularization in PINNs} \label{app:cres}

\begin{figure*}[hb]
            \centering
            
            \begin{subfigure}[b]{0.3\textwidth}
                \centering
                \includegraphics[width=\textwidth]{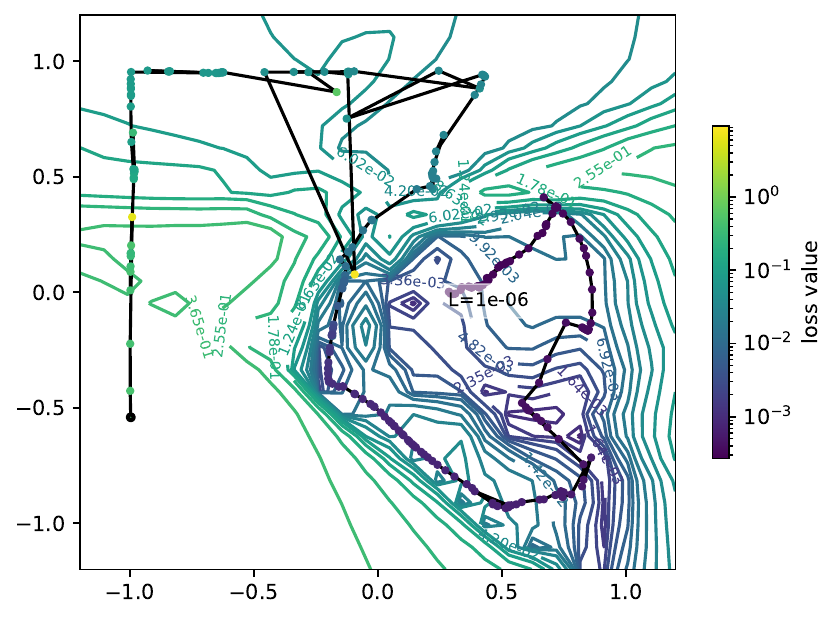}
                \caption{$L=10^{-6}$}
            \end{subfigure}
            \hfill
            \begin{subfigure}[b]{0.3\textwidth}
                \centering
                \includegraphics[width=\textwidth]{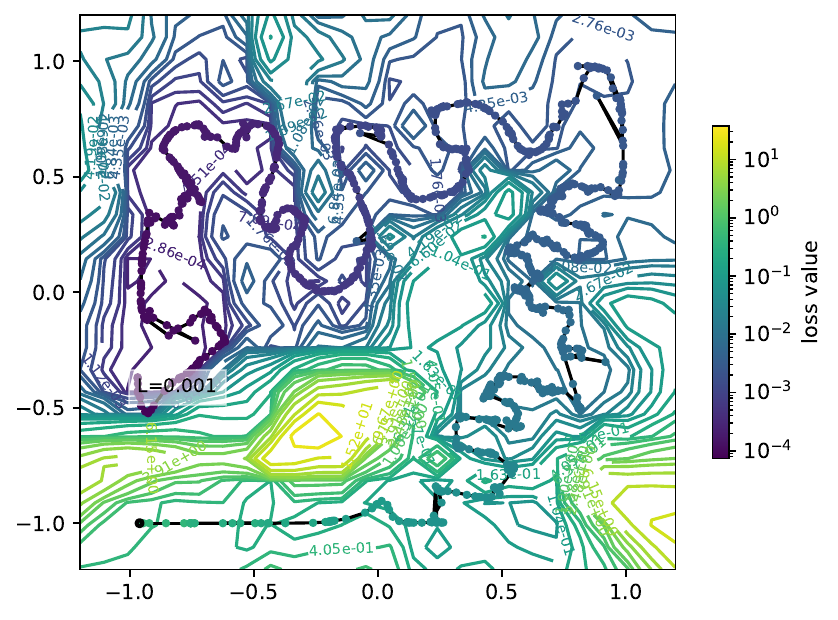}
                \caption{$L=10^{-3}$}
            \end{subfigure}
            \hfill
            \begin{subfigure}[b]{0.3\textwidth}
                \centering
                \includegraphics[width=\textwidth]{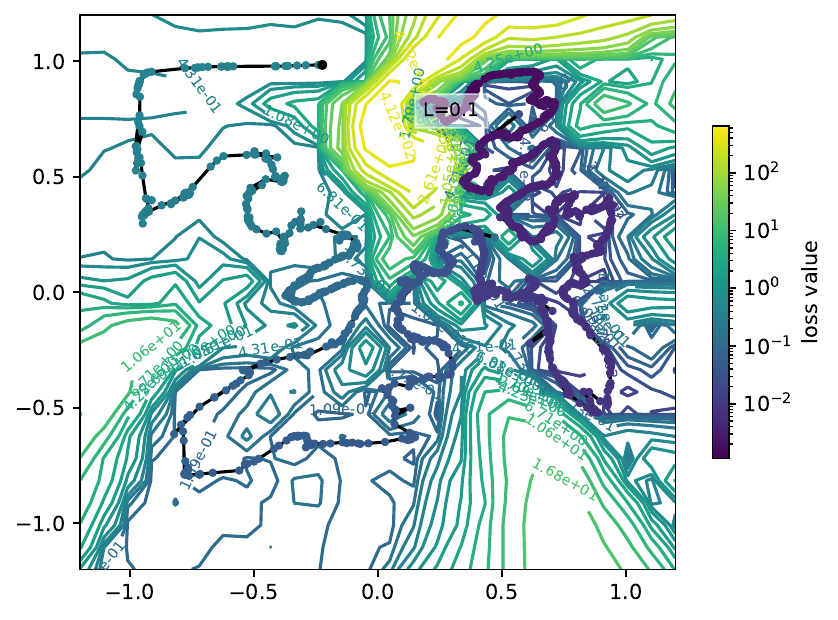}
                \caption{$L=10^{-1}$}
            \end{subfigure}
            
            \caption{A comparison of convection-PDE-solving PINNs with different degrees of soft regularization. \proposedautencoder{} verifies the authors' claim \cite{krishnapriyan2021characterizing} that increasing $\cres$ leads to an $L_{total}$'s landscape that is more non-convex and difficult to optimize.}
            \label{fig:FailsL}
        \end{figure*}

    Similar to \cref{fig:FailsBeta} where the pathological effect of increasing $\beta$ is displayed, \cref{fig:FailsL} shows the effect of increasing $\cres$. Clearly, increasing the regularization factor also leads to an increase in loss landscape complexity.  
    
    \section{A PCA Loss Landscape Analysis of Loss Balancing Methods} \label{app:MTLPCA}
\begin{figure}[htb]
            \centering
              \begin{subfigure}[b]{0.3\textwidth}
                \includegraphics[width=\textwidth]{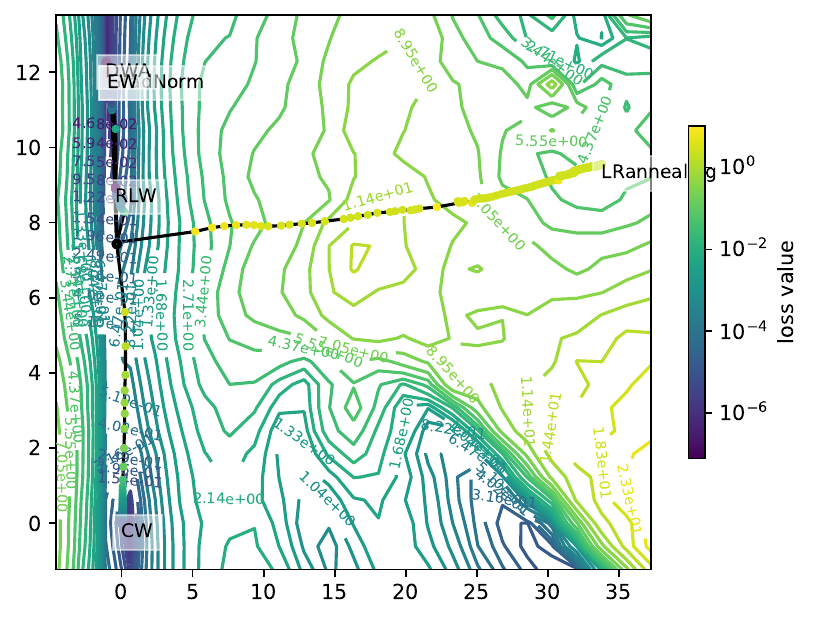}
              \end{subfigure}
                \begin{subfigure}[b]{0.3\textwidth}
                \includegraphics[width=\textwidth]{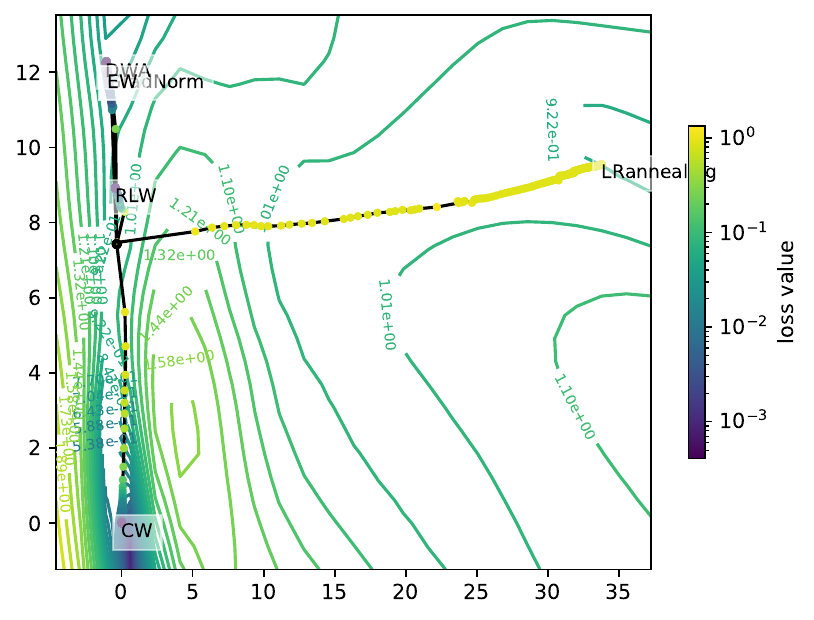}
              \end{subfigure}

              \caption{PCA's loss landscapes of multiple PINN models with different loss balancing methods. Clearly, it is hard to make useful or accurate inferences from this plot compared to the insights found through \cref{fig:dwa}. }
              \label{fig:dwaPCA}
            \end{figure}

         \cref{fig:dwaPCA} shows the PCA loss landscape for the same models in the loss balancing experiment outlined in \cref{sss:MTL} and visualized in \cref{fig:dwa}. As can be seen, PCA fails at delineating whether most models converge at all, converge to the same minima, converge to several minima in the same basin, or different basins altogether.

    \section{Description of Different Loss Balancing Algorithms} \label{app:lossbalancing}

    \cref{tab:MTLmethods} expands on the different loss balancing algorithms that were used in \cref{sss:MTL} and shown in \cref{fig:dwa}.

                \begin{table*}[htb]
            \small
            \centering
            \resizebox{\textwidth}{!}{\begin{tabular}{|p{0.17\textwidth}|p{0.17\textwidth}|p{0.6\textwidth}|}
            \hline
            \textbf{Method} & \textbf{Abbreviation}  & \textbf{Brief description} \\
            \hline
            Equal Weights & \ew{} & All loss terms have a coefficient of $1.0$.\\
            \hdashline
            Constant Weights & \cw{} & Different constants are used to balance the loss terms. In the context of this section, the following values where used based on some hyper-parameter tuning: $\cres=1.0, \cic = \cbc=100.0$.\\
            \hdashline
            Dynamic Weight Averaging & \dwa{} & The weight of each task's loss is adjusted based on the relative improvement of that task's performance compared to the performance of the other tasks.\\
            \hdashline
            Learning Rate Annealing \cite{doi:10.1137/20M1318043} & \lranneal{} & Gradient statistics are utilized during model training to balance the interplay between the losses.\\
            \hdashline
            Gradient Normalization \cite{pmlr-v80-chen18a}& \gradnorm{} & Normalizes the gradients across tasks so that they have similar scales, encouraging the model to focus on the tasks with the most informative gradients. \\
            \hdashline
            Random Loss Weighting \cite{lin2022reasonable}& \rlw{} & The weight of each task is randomly updated each epoch.\\

            \hline
            \end{tabular}}
            
            \caption{A list of the loss balancing algorithms considered in \cref{sss:MTL}. More details on each algorithm can be found in \cite{liang2020simple}. }
            \label{tab:MTLmethods}
            \end{table*}

\end{document}